\documentclass[letter,11pt]{article}
\usepackage[a4paper,top=3cm,bottom=2cm,left=3cm,right=3cm,marginparwidth=1.75cm]{geometry}

\usepackage[english]{babel}
\usepackage[utf8]{inputenc} 
\usepackage[T1]{fontenc}    
\usepackage[colorlinks,allcolors=blue]{hyperref}
\makeatletter
\newcommand{\specificthanks}[1]{\@fnsymbol{#1}}
\makeatother
\usepackage{lmodern}
\usepackage{url}            
\usepackage{booktabs}       
\usepackage{amsfonts}       
\usepackage{nicefrac}       
\usepackage{microtype}      
\usepackage{enumitem}

\usepackage{amsmath}
\usepackage{amsthm}
\usepackage{bm}
\usepackage{bbm}
\usepackage{amsfonts}
\usepackage{amssymb}
\usepackage{mathtools}
\usepackage[ruled, vlined, linesnumbered]{algorithm2e}
\usepackage{graphicx}
\usepackage{subcaption}
\usepackage{subfloat}
\usepackage{natbib}  
\usepackage{caption} 
\usepackage{color}
\usepackage{pgfplots}
\DeclareUnicodeCharacter{2212}{−}
\usepgfplotslibrary{groupplots,dateplot}
\usetikzlibrary{patterns,shapes.arrows}
\pgfplotsset{compat=newest}

\usepackage{appendix}

\newtheorem{theorem}{Theorem}

\newtheorem{definition}[theorem]{Definition}

\newcommand\margin{\mathsf{margin}}

\title{Are Adversarial Examples Created Equal? \\A Learnable Weighted Minimax Risk for Robustness under Non-uniform Attacks}
\author {
    Huimin Zeng \thanks{Equal Contributions}~\thanks{Technical University of Munich. {\tt huimin.zeng@tum.de}}  \and
    Chen Zhu \textsuperscript{\specificthanks{1}}\thanks{University of Maryland, College Park. {\tt \{chenzhu,tomg,furongh\}@cs.umd.edu} } \and
    Tom Goldstein  \textsuperscript{\specificthanks{3}}  \and
    Furong Huang \textsuperscript{\specificthanks{3}} 
}
\date{} 
\begin{document}
\maketitle

\begin{abstract}
Adversarial Training is proved to be an efficient method to defend against adversarial examples, being one of the few defenses that withstand strong attacks. However, traditional defense mechanisms assume a uniform attack over the examples according to the underlying data distribution, which is apparently unrealistic as the attacker could choose to focus on more vulnerable examples. We present a weighted minimax risk optimization that defends against non-uniform attacks, achieving robustness against adversarial examples under perturbed test data distributions. Our modified risk considers importance weights of different adversarial examples and focuses adaptively on harder examples that are wrongly classified or at higher risk of being classified incorrectly. The designed risk allows the training process to learn a strong defense through optimizing the importance weights. The experiments show that our model significantly improves state-of-the-art adversarial accuracy under non-uniform attacks without a significant drop under uniform attacks.
\end{abstract}

\section{Introduction}
\noindent It is widely known that deep neural networks could be vulnerable to adversarially perturbed input examples~\citep{szegedy2013intriguing,huang2017adversarial}.  Having strong defenses against such attacks is of value, especially in high-stakes applications such as autonomous driving and financial credit/risk analysis. 
Adversarial defenses aim to learn a classifier that performs well on both the ``clean'' input examples (accuracy) and the adversarial examples (robustness)~\citep{zhang2019theoretically}. 
Despite a large literature on studying adversarial defenses in machine learning, computer vision, natural language processing and more,
one of the few defenses against adversarial attacks that withstands strong attacks is \emph{adversarial training} \citep{carlini2017adversarial,Kannan2018AdversarialLP, kurakin2016adversarial,  shaham2018understanding}.
In adversarial training, adversarial examples generated via a chosen attack algorithm are included in the training on the fly. 
As is shown in many works \citep{carlini2017adversarial, Kannan2018AdversarialLP, kurakin2016adversarial,shafahi2019adversarial,  shaham2018understanding, zhang2019propagate, zhang2019theoretically}, adversarial training has demonstrated great success in the attack-defense game.

A major issue with adversarial training is that it seeks a model that is robust to adversarial perturbations on the training set. 
Adversarial training attempts to solve a robust optimization problem against a point-wise adversary that independently perturbs each example~\citep{staib2017distributionally}. 
The traditional optimization objective is usually (unweighted) average of robust losses over all training data points; the robust loss for each training data point is evaluated on adversarial example that is independently generated for each training data point
\begin{equation}\label{eq:robust-error}
    \mathbb{E}_{(\bm{x},y) \sim \mathcal{D}_n} \Big[\mathsf{robust\ loss}(f,\bm{x},y,\epsilon)\Big]
\end{equation}
where $\mathcal{D}_n$ is the empirical distribution and the $\mathsf{robust\ loss}(f,\bm{x},y,\epsilon)$ could be any loss function that characterizes the risk of mis-classification of adversarial examples under the threat model of bounded $\epsilon$ perturbation on the input $(\bm{x},y)$ to $f$ (For instance, the 0-1 robust loss is $\bm{1}\{\exists \|\bm{\delta}\|\le \epsilon, \text{s.t. } f(\bm{x}+\bm{\delta})y\le 0\}$).

\begin{figure}[t]
  \centering
  \includegraphics[width=.5\linewidth]{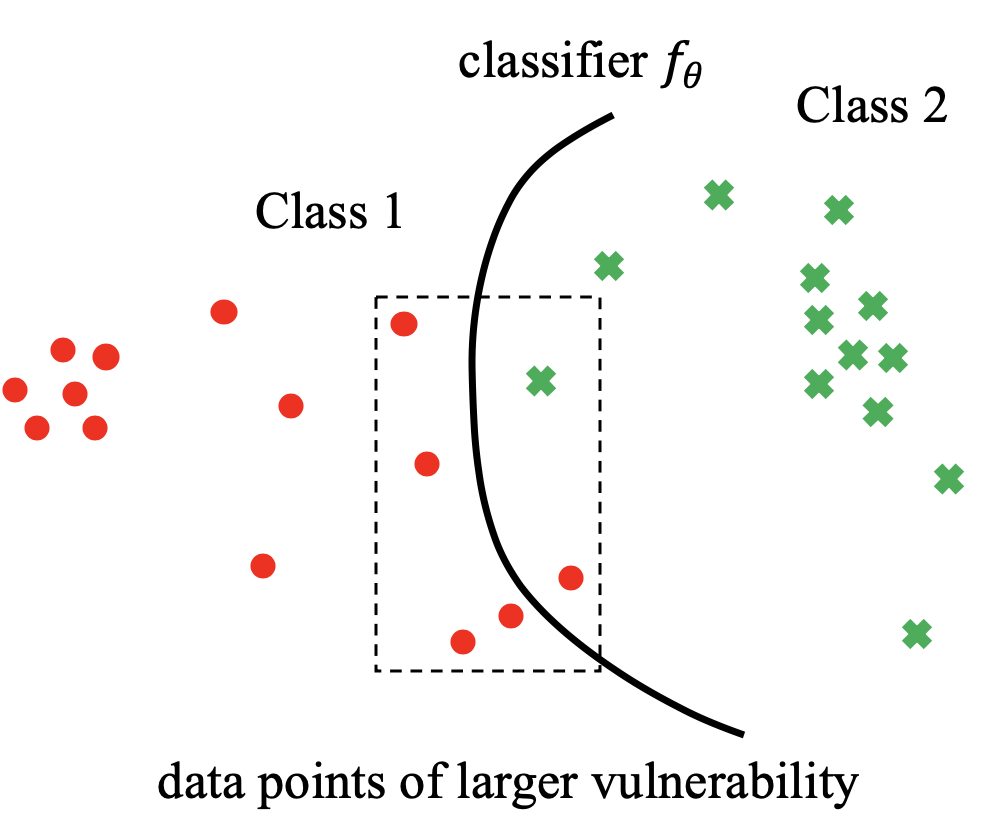}
  \label{fig:sub-first-two}
\vspace{-1em}
\caption{Vulnerability and Distance to Decision Boundary}
\label{fig:vulnerability}
\vspace{-1em}
\end{figure}

This robust error in Equation~\eqref{eq:robust-error} treats the adversarial examples generated around different training data points as equally important when optimizing the training objective. In other words, the training objective assumes that an attacker chooses to attack the input examples uniformly, regardless of how close these examples are to the decision boundary. As a result, the above robust error would fail to measure security against an attacker who focuses on the more vulnerable examples. As shown in Figure~\ref{fig:vulnerability}, the data points that are closer to decision boundary, are more vulnerable to attacks, since the attacker needs a relatively smaller perturbation to move them to wrong side of the decision boundary. 
Therefore, we aim to design robust neural networks against \emph{non-uniform attacks}.

\paragraph{Our methodology.} Motivated by the idea that not all adversarial examples are equally important, we propose a novel weighted minimax risk for adversarial training that achieves both robustness against adversarial examples and accuracy for clean data examples. 
Our modified risk considers importance weights of different adversarial examples and adaptively focuses on vulnerable examples that are wrongly classified or at high risk of being classified incorrectly. The designed weighted risk allows the training process to learn the distribution of the adversarial examples conditioned on a neural network model $\theta$
 through optimization of the importance weights and learn to defend against strong non-uniform attacks.

\paragraph{Summary of Contributions.} 
\begin{enumerate}
    \item We introduce a novel distribution-aware training objective by integrating a re-weighting mechanism to the traditional minimax risk of adversarial training framework. 
    
    \item Based on the distribution-aware minimax risk, we are able to generate stronger adversarial examples, such that some state-of-the-art adversarial training algorithms (for instance, TRADES \citep{zhang2019theoretically}) will perform poorly. On CIFAR10, the robust accuracy of the network (ResNet18 \citep{He2016ResNet}) trained with standard adversarial training setting drops from 53.38\% to 19.78\% under our proposed attacks.

    \item Thirdly, we propose a strong defense mechanism based on our re-weighting strategy, consistently increasing the robustness of models against strong non-uniform (distribution-aware) attacks. Our method improves the state-of-the-art robust accuracy from 19.78\% to 23.62\% on CIFAR10.
    
    \item Besides, our defense mechanism matches the state-of-the-art under traditional evaluation metrics (uniform attacks). On CIFAR10, the network trained with our modified risk is able to achieve 54.10\%, in comparison to the baseline of 53.38\%. 
    
    \item  Finally, we propose two new metrics to evaluate the robustness of the trained classifier under vulnerability- and distribution-aware attacks.
    
\end{enumerate}

\section{Related Work}\label{sec:relatedworks}
A number of defense mechanisms have been proposed to maintain accuracy for adversarial images. This includes detecting and rejecting adversarial examples \citep{ma2018characterizing, meng2017magnet, xu2017feature}, along with other works such as label smoothing and logit squeezing~\citep{mosbach2018logit,shafahi2019label,mosbach2018logit}, gradient regularization~\citep{elsayed2018large,finlay2019scaleable,ross2018improving}, local linearity regularization~\citep{qin2019adversarial}, and a Jacobian regularization~\citep{jakubovitz2018improving}.
Adversarial training proposed by~\citet{madry2017towards} is among the few that are resistant to attacks by~\cite{athalye2018obfuscated}, which broke a suite of defenses. 
Adversarial training defends against test time adversarial examples by augmenting each minibatch of training data with adversarial examples during training. 

Adversarial training is powerful in terms of defending against adversarial examples. We witnessed a surge of studies on designing loss functions for training robust classifiers. Many methods in the adversarial training literature treat all training examples equally without using sample-level information. 

Recently, however, \citet{balaji2019instance} propose example-specific perturbation radius around every training example to combat the adversarial training's failure to generalize well to unperturbed test set. 
Moreover, \citet{zhang2019theoretically} provides a theoretical characterization of the trade-off between the natural accuracy and robust accuracy by investigating the Bayes decision boundary and introducing a new regularization based on the KL divergence of adversarial logit pairs, with which the trained model reaches state-of-the-art performance. 
While adversarial training improves robustness at the cost of generalization on clean samples for image classification, recent works have shown it is possible to improve generalization on clean samples for language Transformers and graph neural networks~\citep{zhu2019freelb,jiang2019smart,gan2020large,kong2020flag}.

\emph{Distributionally robust optimization} (DRO) is a tool that links generalization and robustness~\citep{staib2017distributionally,ben2013robust,blanchet2017doubly,delage2010distributionally,duchi2016statistics,gao2016distributionally,goh2010distributionally}. DRO seeks a model that performs well under adversarial joint perturbations of the entire training set. The adversary is not limited to moving points individually, but can move the entire distribution within an $\epsilon$-ball of $\mathcal{D}_n$ for some notion of distance between distributions. The attacker has a specific attack budget to attack the distribution of the dataset; the perturbed distribution has to be $\epsilon$-close to the uniform distribution. However in the non-uniform attack setting we consider, although the attacker might have constrained power to alter each image, their attack to the distribution might be unconstrained. 

\section{Weighted Minimax Risk Models}\label{sec:weighted-risk}
\subsection{Rethinking Adversarial Training}\label{subsec:rethinking}
\noindent \textbf{Traditional training}
Traditional model training is the process of learning optimal model parameter $\bm{\theta}$ that characterizes a mapping  from input space to output space $f_{\bm{\theta}}:\bm{\mathcal{X}} \to \mathcal{Y}$. 
The model is designed to minimize the expectation of the \emph{natural loss function} $l(f_{\bm{\theta}}(\bm{x_{i}}), y_{i})$ under the unknown underlying distribution of input examples $(\bm{x}_i,y_i) \sim \mathcal{D}$
\begin{equation}\label{eq:traditional-training}
    \min\limits_{{\bm{\theta}}}
    \mathbb{E}_{(\bm{x}_i,y_i)\sim \mathcal{D}}
 \Big[l(f_{\bm{\theta}}(\bm{x_{i}}), y_{i})\Big]
\end{equation}

In practice, an assumption of input examples $(\bm{x}_i,y_i)_{i=1}^N$ being i.i.d. is often made, allowing unbiased empirical estimation of the expectation of the \emph{natural} loss. 
\begin{equation}\label{eq:traditional-training-empirical}
    \min\limits_{{\bm{\theta}}}\frac{1}{N}
    \sum\limits_{i=1}^{N}
 l(f_{\bm{\theta}}(\bm{x_{i}}), y_{i})
\end{equation}

Performing full-batch gradient descent is too computationally expensive. Therefore, the models are usually trained by means of mini-batch gradient descent with batch size $m$. This is important, since this is statistically equivalent to full-batch gradient descent, but with larger variance, which is related to batch size.
\begin{equation}\label{eq:traditional-training-empirical_minibatch}
    \min\limits_{{\bm{\theta}}}\frac{1}{m}
    \sum\limits_{i=1}^{m}
 l(f_{\bm{\theta}}(\bm{x_{i}}), y_{i})
\end{equation}

\noindent \textbf{Adversarial training} Adversarial training has been one of the most prevalent approaches to combat evasion attacks. 
Specifically, adversarial training solves a mini-max problem by alternating between a network parameter update and an update on input perturbations using projected stochastic gradient descent, seeking a convergence to an equilibrium. 
The optimizer, originally designed to minimize the natural loss on clean data examples, now takes additional adversarial examples generated during training into consideration. 
On each step, an inner loop generates the strongest perturbation $\bm{\delta}_i$ within the $\epsilon$ radius of each input example $\bm{x}_i$ (a specific norm bounded by $\epsilon$~\citep{szegedy2013intriguing}) using projected gradient descent (PGD), and then minimizes the \emph{adversarial loss function} $l(f_{\bm{\theta}}(\bm{x_{i}}+\bm{\delta_{i}}), y_{i})$ in expectation according to distribution $\mathcal{D}$
\begin{equation}\label{eq:adversarial-training}
    \min_{{\bm{\theta}}}\mathbb{E}_{(\bm{x}_i,y_i)\sim \mathcal{D}}\Bigg[
    \max_{\bm{\delta_{i}}:  \| \bm{\delta_{i}} \|<\epsilon} l(f_{\bm{\theta}}(\bm{x_{i}}+\bm{\delta_{i}}), y_{i}).
    \Bigg]
\end{equation}
Therefore, during each update of the network parameters, adversarial examples (perturbations of the input examples) are generated through PGD search of a perturbation direction that maximizes the loss function, and are added to the input examples for next update of the network parameters. 
The idea behind adversarial training is that these adversarially generated perturbations, added to the training data, will force the model to proactively adjust the model parameters during training to combat potential adversarial perturbations at test time.

Corresponding to Equation~\eqref{eq:traditional-training-empirical_minibatch}, where the optimization objective is constructed over mini-batches, the assumption of adversarial examples being i.i.d. is still made for unbiased empirical estimation of the expectation of the \emph{adversarial} loss 
\begin{equation}\label{eq:adversarial-training-empirical}
    \min_{{\bm{\theta}}}\frac{1}{m}
    \sum\limits_{i=1}^{m}
    \max_{\bm{\delta_{i}}: \| \bm{\delta_{i}} \|<\epsilon} l(f_{\bm{\theta}}(\bm{x_{i}}+\bm{\delta_{i}}), y_{i})
\end{equation}

\noindent \textbf{Are adversarial examples created equal?}
In traditional training, it is reasonable to use the non-weighted sum of the loss evaluated at each data point as an unbiased estimation of the expectation of the natural loss. 
However in adversarial training, one often ignored issue is that the loss we optimize is no longer the natural loss on clean data. 
The goal of adversarial training is to combat adversarial examples at test time. 
Robustness is achieved by generating a strong (if not the strongest) adversarial perturbation $\bm{\delta}_i$ for each training data point $(\bm{x}_i,y_i)$. However it is unclear whether we should treat the generated adversarial examples $\{\bm{\delta}_i\}_{i=1}^N$ equally.
In particular, the loss function in Equation~\eqref{eq:adversarial-training-empirical} suffers from two problems. \\
\textbf{problem (a):}
It puts equal weights on adversarial examples closer to the decision boundary and examples far away;\\
\textbf{problem (b):}
It assumes that a white-box attacker will always perform a uniform attack on all data points, but, in practice, it might attack the distribution of the adversarial example as well.

In the following section, we will introduce a modified adversarial loss, called \emph{weighted minimax risk}, where the weights are learnable via a training process. 
We focus on $\ell_\infty$ norm bounded perturbations although the mechanism could be extended to other norms.


\subsection{Re-weighting of  Vulnerability and Robustness}
In Section~\ref{subsec:rethinking}, \textbf{problem (a)} points to a potential problem with Equation~\eqref{eq:adversarial-training-empirical} --- all adversarial examples generated during adversarial training, despite their varying distances to the decision boundary and thus varying risk of being misclassified, are treated equally when empirically estimating the expectation of the adversarial loss. 

\textbf{Problem (b)}, on the other hand, reveals another unsatisfactory design of Equation~\eqref{eq:adversarial-training-empirical} --- due to the adversarial nature of evasion attacks, the test time adversarial examples $\bm{x}_{\text{test}}'$ 
do not necessarily have the same distribution as the training time adversarial examples $\bm{x}_{\text{training}}'$ generated in adversarial training. It is highly likely that the distribution of the adversarial risk is not equal to the independent identical distribution of clean data points. 

In this subsection, we first define the ``confidence margin'' as a measurement of vulnerability of examples in the probability space. Positive margin indicates a correctly classified example and negative margin an incorrectly classified one.

\begin{definition}[\bf margin of a classifier $f$ on example $(\bm{x}_i,y_i)$ ~\citep{zhang2019defending}]\label{def:margin} For a data point $(\bm{x}_i,y_i)$, the margin is the difference between the classifier's confidence in the correct label $y_i$ and the maximal probability of an incorrect label $t$, 
$\margin(f,\bm{x}_i,y_i) = p(f(\bm{x}_i)=y_i) - \max_{t \neq y_i}p(f(\bm{x}_i)=t)$.
\end{definition}

\noindent \textbf{Remark} In the context of white-box attack, this margin is unfortunately accessible to the adversarial attackers. This is the key prerequisite for an adversarial attacker to perform non-uniform attack (more details in Section~\ref{sec:attacking_distribution}).

Although it is impossible to know the distribution of test time adversarial examples, we could follow a principle to reduce the vulnerability of our model by focusing on vulnerable examples. 
In particular, we aim to design an importance weight $c_i$ based on the margin of $\bm{x}_{\text{training}}'$. If the margin of the generated adversarial example during training $\bm{x}_{\text{training}}'$ is large, the adversarial example $\bm{x}_{\text{training}}'$ is a weak attack (a positive margin indicates the attack failed), and thus its importance weight $c_i$ should be smaller. A more detailed description follows below.

\begin{enumerate}
 \item if margin is \textbf{positive} and \text{large} (the adversarial $\bm{x}_{\text{training}}'$ is \textbf{correctly} classified and rather robust), the importance weight $c_i$ should be \textbf{small};
    \item If margin is \textbf{positive} but \textbf{small} (the adversarial $\bm{x}_{\text{training}}'$ is \textbf{correctly} classified but \textbf{vulnerable}), the importance weight  $c_i$ should be \textbf{moderate};
    \item if margin is \textbf{negative} (the adversarial $\bm{x}_{\text{training}}'$ is \textbf{incorrectly} classified), the importance weight  $c_i$ should be \textbf{large}.
\end{enumerate} 

\begin{figure}[t]
  \centering
  \includegraphics[width=.6\linewidth]{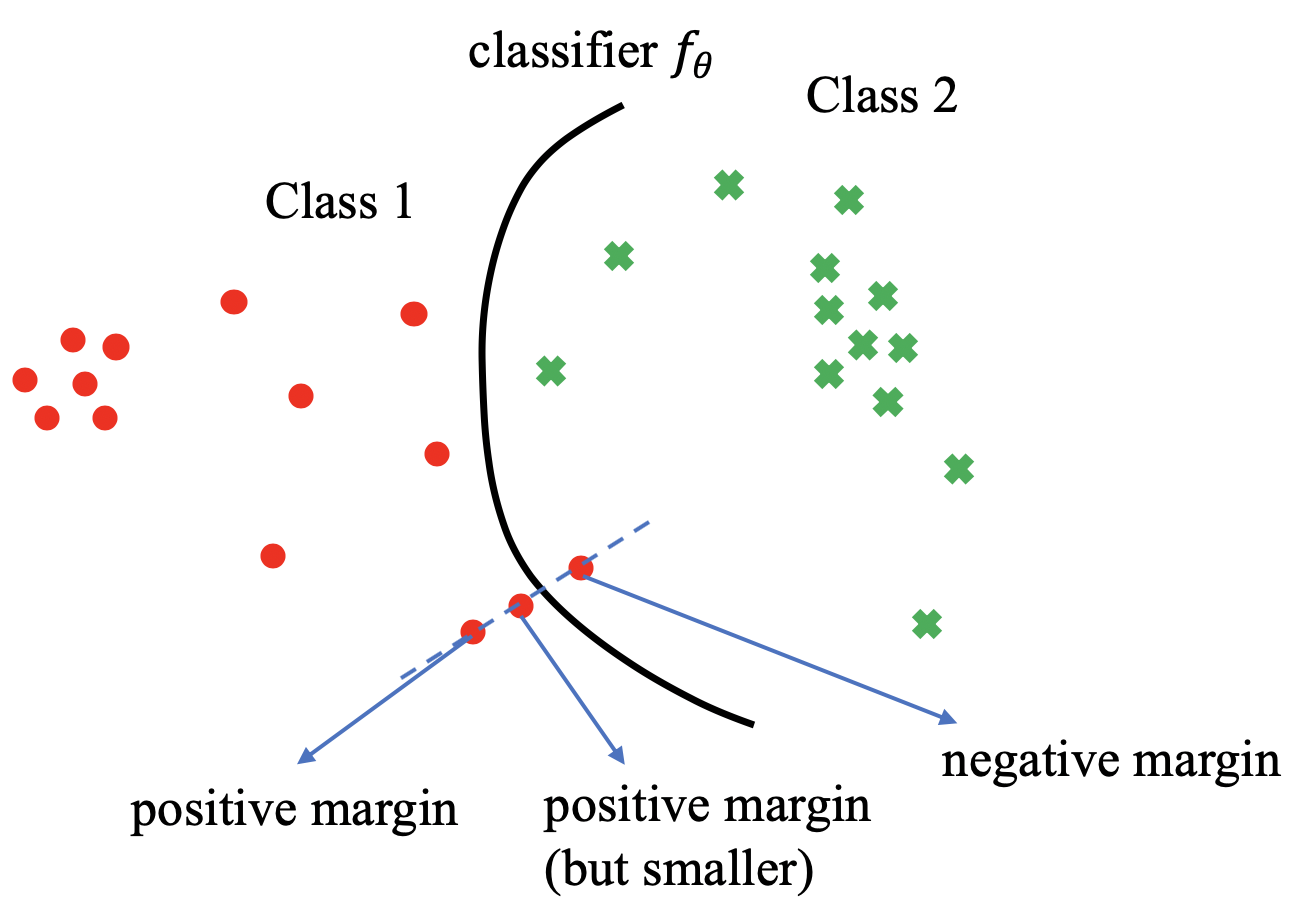}
  \label{fig:sub-first-one}
\vspace{-1em}
\caption{\small Margin and Vulnerability}
\label{fig:margin_vulnerability}
\vspace{-1em}
\end{figure}
Figure~\ref{fig:margin_vulnerability} shows the relation between margin and vulnerability of certain data points. It is straightforward to design a loss function, so that the focus of training is on the examples which are easier to be attacked (corresponding to small positive margin) or are already successfully attacked (corresponding to negative margin). Now, we formally propose Adaptive Margin-aware Risk.
\paragraph{Adaptive Margin-aware Risk} \emph{Adaptive margin-aware minimax risk} is a minimax optimization objective, using an exponential family parameterized by the margin of the adversarial examples in training. 
{\small\begin{equation}
    \label{eq:adversarial-training-margin}
    \min_{{\bm{\theta}}}
    \sum\limits_{i=1}^{m}
    \max_{\bm{\delta_{i}}: \| \bm{\delta_{i}} \|<\epsilon} 
    e^{-\alpha \ \margin(f_{\bm{\theta}},\bm{x}_i+\bm{\delta}_i,y_i)}
    \
    l(f_{\bm{\theta}}(\bm{x_{i}}+\bm{\delta_{i}}), y_{i})
\end{equation}}
where $\alpha > 0$ is a positive hyperparameter of this exponential weight kernel. With the intuition, we can see that there is a positive correlation between the exponential weight kernel and individual loss $l$. Larger individual loss will induce a larger weight, and vice versa.

\noindent \textbf{Comparison with ``natural and adversarial loss combined''} Previous works \citep{goodfellow2014explaining, kurakin2016adversarial} consider a loss that combines both the natural loss and adversarial loss with a hyperparameter $\lambda$, i.e., 
\begin{equation}
\label{eq:adversarial-training-empirical-combine}
  \min\limits_{{\bm{\theta}}}
    \underbrace{\sum\limits_{i=1}^{m}
 l(f_{\bm{\theta}}(\bm{x_{i}}), y_{i})}_{\text{natural loss}}
 + \lambda
    \underbrace{\sum\limits_{i=1}^{m}
    \max_{\bm{\delta_{i}}: \| \bm{\delta_{i}} \|<\epsilon} l(f_{\bm{\theta}}(\bm{x_{i}}+\bm{\delta_{i}}), y_{i})}_{\text{adversarial loss}} . 
\end{equation}
This approach could be thought of as a limiting case of our proposed margin kernel with small $\alpha$, and it doesn't account for weighting adversarial examples with varying amplitudes. 

As we see, Equation~\eqref{eq:adversarial-training-empirical-combine} designs a defense mechanism that treats adversarial examples $\bm{x}_{\text{training}}'$ equally and would fail if the attacker at test time chooses to attack the more vulnerable examples (closer to the decision boundary). This is a key difference compared to natural training when unseen examples are assumed to be from the same distribution as the training examples.

\section{Distributionally Robust Adversrial Training}
\label{sec:algo}
\subsection{Attack Distribution of Adversarial Examples}

The distribution of examples that the adversary deploys to attack, i.e., the attack distribution of adversarial examples may deviate from the empirical distribution $\mathcal{D}_n$ represented by the training examples.
In the context of adversarial training, 
the objective function we use to achieving robustness against an ``attack distribution-aware'' adversary should be 
\begin{equation}\label{eq:adversarial-training-correct}
    \mathcal{L}'(\bm{\theta}) =\mathbb{E}_{(\bm{x}'_{},y_{})\sim \mathcal{D}'}\Bigg[
    l(f_{\bm{\theta}}(\bm{x'_{ }}), y_{ })
    \Bigg],
\end{equation}
where $\mathcal{D}'$ denotes the unknown underlying distribution of the adversarial examples. 

In a standard adversarial training framework, as reviewed in Section~\ref{sec:weighted-risk}, the learner generates the perturbation $\bm{\delta}_i^*$ (using PGD) and thus an adversarial example $\bm{x}_i' = \bm{x}_i + \bm{\delta}_i^*$ for each input example $\bm{x}_i$ to minimize the \emph{adversarial loss}. The training objective used in practice is 

{\small\begin{equation}\label{eq:adversarial-training-wrong}
    \widehat{\mathcal{L}}(\bm{\theta}) =\frac{1}{m}\sum\limits_{i=1}^m
     l(f_{\bm{\theta}}(\bm{x_{i}}+\bm{\delta}_i^*), y_{i}) =\frac{
     1}{m}\sum\limits_{i=1}^m
     l(f_{\bm{\theta}}(\bm{x_{i}}'), y_{i})
\end{equation}}

The \textbf{problem} is that the objective $\widehat{\mathcal{L}}(\bm{\theta}) $ (Equation~\eqref{eq:adversarial-training-wrong}) used in standard adversarial training is often \textbf{not} an unbiased estimator of the true objective function $\mathcal{L}'(\bm{\theta})$ (Equation~\eqref{eq:adversarial-training-correct}) required, since the generated adversarial examples during adversarial training $({\bm{x}_i}',y_i)$ are not necessarily good representation of the underlying distribution of the adversarial examples. This is exactly the challenge of achieving robust models; the adversarial attacks are unpredictable. 

The true objective $\mathcal{L}'(\bm{\theta})$ illustrated in Equation~\eqref{eq:adversarial-training-correct} is unfortunately often intractable, since the underlying distribution of the adversarial examples is unknown. 
The problem reduces to an unbiased estimation of the unknown distribution of the adversarial examples.

\noindent \textbf{Comparison with distributionally robust optimization} In distributionally robust optimization (DRO) literature as surveyed in Section~\ref{sec:relatedworks}, the methods developed often assume that the divergence between the empirical distribution and the attack distribution is bounded by a threshold  $\mathsf{divergence}(\mathcal{D}_n,\mathcal{D}')\le \rho$. Thus, the DRO~\citep{namkoong2016stochastic} objective is 
\begin{equation}\label{eq:dro}
    \min\limits_\theta \max\limits_{\mathsf{divergence}(\mathcal{D}_n,\mathcal{D}')\le \rho} \mathbb{E}_{(\bm{x},y)\sim \mathcal{D}'}[l(f_{\bm{\theta}}(\bm{x}),y)]
\end{equation}

Apart from the complexity of solving the inner constrained maximization problem, DRO requires evaluating the loss for every example in the entire training set before every minimization step, which can be expensive for practical models and datasets. 

As illustrated in Section~\ref{sec:weighted-risk}, we introduce a risk estimator for each data point individually, so that the objective function is able to express the distribution of the adversarial examples (allowing a non-uniform attack) and learn it via training. 
It only requires evaluating an importance weight at each sample in the minibatch, but is able to improve distributional robustness against adversarial examples, as we will show in Appendix~\ref{appendix:sensi_hyper}.

\begin{definition}[\bf{Importance Weights}]
\label{def:scaling} For training data points $(\bm{x}_i, y_i)_{i=1}^N$ and their corresponding adversarial perturbations $(\bm{x}_i', y_i)_{i=1}^N$, 
we define the importance weight $s(f_{\bm{\theta}}, \bm{x}_i', y_i)$ between $(\bm{x}_i', y_i)$ and $(\bm{x}_i, y_i)$, i.e., the ratio of the adversarial example distribution and the clean data distribution evaluated at training data point $(\bm{x}_i, y_i)$, as
\begin{equation}
s(f_{\bm{\theta}}, \bm{x}_i', y_i) := \frac{\mathcal{D}'(\bm{x}'_i,y_i)}{\mathcal{D}(\bm{x}_i,y_i)}.
\end{equation}
\end{definition}
\noindent\textbf{Remark} In our adaptive margin-aware risk, the importance weight is parameterized as the learnable scaling factor  $s(f_{\bm{\theta}}, \bm{x}_i', y_i) = e^{-\alpha \ \margin(f_{\bm{\theta}},\bm{x}_i+\bm{\delta}_i,y_i)}$ as shown in Equation~\eqref{eq:adversarial-training-margin}.

Therefore, our re-weighting strategy -- adaptive margin-aware risk-- proposes to train the objective function as follows (if we consider full-batch gradient descent)
\begin{align}
 \widetilde{\mathcal{L}}(\bm{\theta}) 
    \quad  &  = \quad \frac{1}{N}\sum\limits_{i=1}^N
      s(f_{\bm{\theta}}, \bm{x}_i', y_i)
     l(f_{\bm{\theta}}(\bm{x_{i}}'), y_{i}) \label{eq:adversarial-training-ours} \\
     \quad  &  \approx \quad \mathbb{E}_{({\bm{x}}, y)\sim \mathcal{D}}\Big[
      {s(f_{\bm{\theta}}, \bm{x}', y)}
     l(f_{\bm{\theta}}(\bm{x_{}}'), y_{})\Big] \label{eq:adversarial-training-ours-compact-clean}\\
     \quad  &  
  \approx
     \quad \mathbb{E}_{({\bm{x}'}, y)\sim \mathcal{D}'}\Big[
     l(f_{\bm{\theta}}(\bm{x_{}}'), y_{})\Big]. \label{eq:adversarial-training-ours-compact-adversarial}
\end{align}

Since the importance weight scaling factors $s(f_{\bm{\theta}}, \bm{x}', y)$ is learnable, 
our objective can be thought of as ``learning''
the \emph{adversarial example distribution conditioned on a neural network model $\bm{\theta}$} via learning of the importance weight ${s(f_{\bm{\theta}}, \bm{x}', y)}$ using the objective in Equation~\eqref{eq:adversarial-training-ours}. 

Based on the previous analysis of computational feasibility in Section~\ref{sec:weighted-risk}, it is impractical to perform the full batch optimization regarding such problem. However, we verify that minimizing adaptive margin-aware risk in mini-batches is statistically equivalent to a full-batch version.
\begin{equation}
    \label{eq:mini__vs_full}
    \begin{split}
        & \mathbb{E}_{(\bm{x}_i, y_i)\sim \mathcal{D}}[\widetilde{L}_m(\bm{\theta})] \\
        &= \mathbb{E}_{(\bm{x}_i, y_i)\sim \mathcal{D}} [\frac{1}{m} \sum_{i=1}^m s(f_{\bm{\theta}}, \bm{x}_i', y_i) l(f_{\bm{\theta}}(\bm{x}_i'), \bm{y}_i)] \\
     &=\widetilde{L}(\bm{\theta})
    \end{split}
\end{equation}
The proof is shown in Appendix~\ref{appendix:mini}.

\subsection{Defending against vulnerability- and distribution-aware attacks}
\label{sec:attacking_distribution}
As we have argued before, a ``smarter'' white-box attacker could have access to the vulnerability of different adversarial examples, and therefore could focus on more vulnerable examples. More important, the attacker is able to \textbf{sample} the more vulnerable data points more frequently and craft adversarial perturbations to these sampled examples. 
In our work, the vulnerability is measured by the margin-aware weights. If the vulnerability of a data point is larger, then its margin-aware weight is larger, and it will be sampled by the attacker with higher probability. 

To develop an efficient defense mechanism against non-uniform attacks, we augment the adversarial training framework using our proposed adaptive margin-aware risk, as shown in Algorithm~\ref{alg:importance-based_attack}.

\begin{algorithm}
\SetAlgoLined
  \caption{Weighted adversarial training}
    \label{alg:importance-based_attack}
    \textbf{Inputs} 
    network $f_{\bm{\theta}}$, training examples $\{ \bm{x}_i \} _{i=1}^{N}$, number of steps for PGD $K$ and step size of PGD $\eta_1$, learning rate $\eta_2$\;
    \textbf{Output}: robust network $f_{\bm{\theta}^*}$;
    
    \For{training iterations}{
         Read a mini-batch $B =\{ \bm{x}_1, \bm{x}_2, ..., \bm{x}_m \}$ from training set \;
         \For{i=1,..,m}
         {
            Initialize $\bm{x'}_i = \bm{x}_i + 0.001 \bm{\xi}$,
             where $\bm{\xi} \sim \mathcal{N}(\bm{0}, \bm{I})$ \;
            \For{k = 1, 2, ..., K}
            {
                $\mathcal{L}(\bm{x'}_i) = s(f_{\bm{\theta}^*}, \bm{x'}_i, y_i)l(f_{\bm{\theta}^*}(\bm{x'}_i), y_i)$, where $s(f_{\bm{\theta}}, \bm{x}_i', y_i) = e^{-\alpha \ \margin(f_{\bm{\theta}},\bm{x}_i+\bm{\delta}_i,y_i)}$ \;
                $\bm{x'}_i = \prod_{\mathbb{B}(\bm{x}_i, \epsilon)}(\bm{x'}_i + \eta_1 \textsf{sign} \nabla_{\bm{x}_i}\mathcal{L}(\bm{x'}_i))$, where $\prod$ is the projection operator\;
            }
    }
    $\theta = \theta - \eta_2 \nabla_{\bm{\theta}}\mathcal{L}(\bm{x'}_i))$ 
    }
\end{algorithm}

\noindent \textbf{Evaluation} During evaluation, for any test example $(\bm{x}_i, y_i)$, we define the normalized importance weights (normalized margin-aware weights) $\tilde{s}(f_{\bm{\theta}}, \bm{x}_i', y_i)$ as $ \tilde{s}(f_{\bm{\theta}}, \bm{x}_i', y_i):=\frac{s(f_{\bm{\theta}}, \bm{x}_i', y_i)}{\sum_{i=1}^{N_{\mathsf{test}}}s(f_{\bm{\theta}}, \bm{x}_i', y_i)}$.
The normalized margin-aware weights could be interpreted as the the probability of attacking example $(\bm{x}_i, y_i)$. 
A uniform attack implies that the probability of attacking example $(\bm{x}_i, y_i)$ is  $\frac{1}{N_{\mathsf{test}}}$. 
For a non-uniform attack, the probability of attacking example $(\bm{x}_i, y_i)$ is $\tilde{s}(f_{\bm{\theta}}, \bm{x}_i', y_i)$. 

We argue that the traditional evaluation under uniform attack should be improved under the setting of non-uniform attack. More details are in the next section.

\section{Experiments}\label{sec:experiments}
\subsection{Evaluation metrics}
\noindent \textbf{Traditional evaluation metrics} Traditionally, we measure the performance of each method using natural accuracy on clean data, denoted as $\mathcal{A}_{nat}$.
Robust accuracy $\mathcal{A}_{rob}$ is commonly used to evaluate the adversarial accuracy
\begin{equation}
    \mathcal{A}_{rob} =
    \frac{1}{N_{\mathsf{test}}}\sum_{i=1}^{N_{\mathsf{test}}}[\mathbbm{1}(f_{\bm{\theta}^*}(\bm{x}_i+\bm{\delta}_i^*), y_i)]
\end{equation}
on the test examples uniformly. Note that $\mathcal{D}_n^{\text{test}}$ is the empirical distribution of the clean test examples and $\bm{\delta}^*$ is derived using the traditional unweighted loss function. 

\noindent \textbf{Our evaluation metric I: $\mathcal{A}_{sa}$} As we motivate in this paper, the evaluation of $\mathcal{A}_{rob}$ makes an unrealistic assumption that the adversary chooses to attack uniformly (although the perturbations at different examples $\bm{\delta}_i^*$ are different). Therefore $\mathcal{A}_{rob}$ is not necessarily the best way to evaluate the performance of the robustness under non-uniform attacks. We introduce an modified accuracy, namely $\mathcal{A}_{sa}$ that evaluate robustness under non-uniform attacks. To compute $\mathcal{A}_{sa}$, the perturbations ${\bm{\delta'}}_i^*$ are crafted independently using the traditional unweighted loss. However, the attacker attacks the test examples non-uniformly, i.e., the adversarial examples are \textbf{sampled} according to a non-uniform distribution --- the normalized importance weights:
\begin{equation}
    \mathcal{A}_{sa} =
    \frac{1}{N_{\mathsf{test}}}\sum_{i=1}^{N_{\mathsf{test}}}
    [\tilde{s}(f_{\bm{\theta}^*}, \bm{x}_i+{\bm{\delta'}}_i^*, y_i)
    \mathbbm{1}(f_{\bm{\theta}^*}(\bm{x}_i+{\bm{\delta'}}_i^*), y_i)]
\end{equation}

\label{sec:metrics}
\noindent \textbf{Our evaluation metric II: $\mathcal{A}_{tr}$}Furthermore, we propose another evaluation metric, $\mathcal{A}_{tr}$. Here, the adversarial examples are not only crafted with importance-weighted loss, but also under the sophisticated selection (importance-based sampling):
\begin{equation}
    \mathcal{A}_{tr} = 
    \frac{1}{N_{\mathsf{test}}}\sum_{i=1}^{N_{\mathsf{test}}}
     [\tilde{s}(f_{\bm{\theta}^*}, \bm{x}_i+{\bm{\delta''}}_i^*, y_i)
    \mathbbm{1}(f_{\bm{\theta}^*}(\bm{x}_i+{\bm{\delta''}}_i^*), y_i)]
\end{equation}

The perturbations   $\bm{\delta''}^*$ are generated via the process in Algorithm~\ref{alg:importance-based_attack}.
This $\mathcal{A}_{tr}$ reflects to what extent the attacker is able to transfer the margin-aware weights into the efficacy of the adversarial attacks, in terms of the generative process as well as  sampling process.

\noindent \textbf{Remark} Empirically, these three metrics correspond to three different kind of adversarial attackers of different attacking power. $\mathcal{A}_{rob}$ is the traditional robust accuracy. Regarding this accuracy, the adversary is the weakest one in comparison to the others. This ``naive'' attacker attacks all samples uniformly and does not leverage the vulnerability of individual data points. $\mathcal{A}_{sa}$ is the accuracy evaluated on the adversarial examples, which are generated by the \textbf{unweighted} loss but sampled non-uniformly based on the normalized importance weights. When computing $\mathcal{A}_{sa}$, the network is dealing with a smarter attacker, since the adversary knows to attack vulnerable examples more frequently. Finally, $\mathcal{A}_{tr}$ measures the robustness of the trained model in the hardest case, where the adversarial examples are generated based on the \textbf{weighted} loss, but also are sampled based on the normalized importance weights. In this case, the attacker is the strongest one. It assigns larger energy to attack more vulnerable examples and samples such vulnerable adversarial examples more frequently. Therefore, when all hyperparameters ($\epsilon$, $\alpha$) are the same, we expect $\mathcal{A}_{tr} \leq \mathcal{A}_{sa} \leq \mathcal{A}_{rob}$ in most scenarios.

\noindent \textbf{Hyperparameters of the margin-aware weights} Recall that the importance weight $s(f_{\bm{\theta}}, \bm{x}_i', y_i) = e^{-\alpha \ \margin(f_{\bm{\theta}},\bm{x}_i+\bm{\delta}_i,y_i)}$.
For a better understanding of the results, we clarify that the $\alpha$ used during training will be denoted as $\alpha_{train}$. During test, the non-uniform attack model used to evaluate the robustness of a trained network uses the importance weight parameterized by $\alpha_{test}$. Regardless training or testing, the value of $\alpha$ indicates the power of the adversarial attacker. If $\alpha_{train}$ is larger, then a stronger non-uniform attacker is included during training. Therefore, the resulted model should be able to withstand stronger non-uniform attacks. Similarly, if $\alpha_{test}$ is large, the attacker is able to exaggerate the re-weighting effect to a larger extent, corresponding to stronger attack power. 

\begin{table}[!htbp]
\centering
\caption{\small \textbf{Robustness against non-uniform attacks on MNIST (Proposed Metrics).}
The adversarial examples are generated through 40-PGD with $\epsilon=0.3$.}  
\label{tab:adv_eps_compare_trades_tr_mnist}
\vspace{-0.5em}
\resizebox{0.6\columnwidth}{!}{\begin{tabular}{c|c|c|c|c|c}
\hline
\textbf{Defense} & $\alpha_{train}$ &$\alpha_{test}$&  $\mathcal{A}_{\text{rob}}$ (\%) & $\mathcal{A}_{\text{sa}}$ (\%) &$\mathcal{A}_{\text{tr}}$(\%)  \\
\hline
\hline
 PGD & - & 1.0 & 93.95 & 74.85 & 74.65 \\
 PGD+ours & 0.5 & 1.0 & 95.22  & \textbf{80.54} & \textbf{80.53} \\
\hline
 PGD & - & 1.5 & 93.95 & 56.10 & 55.87 \\
 PGD+ours & 0.5 & 1.5 & 95.22  & \textbf{64.89} & \textbf{64.63} \\
\hline
 PGD & - & 2.0 & 93.95 & 35.32 & 35.04 \\
 PGD+ours & 0.5 & 2.0 & 95.22  & \textbf{44.96} & \textbf{44.70} \\
\hline
\hline
 TRADES & - & 1.0 & 95.59 & 83.18 & 83.07 \\
 TRADES+ours & 2.0 & 1.0 & 95.20 & \textbf{86.34} & \textbf{85.94} \\
 \hline
 TRADES & - & 1.5 & {95.59} & 70.07 & 69.72 \\
 TRADES+ours & 2.0 & 1.5 & 95.20 & \textbf{78.10} & \textbf{77.22} \\
 \hline
 TRADES & - & 2.0 & {95.59} & 52.15 & 51.52 \\
 TRADES+ours & 2.0 & 2.0 & 95.20 & \textbf{66.71} & \textbf{65.61} \\
 \hline
\end{tabular}}
\end{table}

\begin{table}[!htbp]
\centering
\caption{\small \textbf{Robustness against non-uniform attacks on CIFAR10 (Proposed Metrics).} 
The adversarial examples are generated through 20-PGD with $\epsilon=0.031$.}  
\label{tab:adv_eps_compare_trades_tr_cifar10}
\vspace{-0.5em}
\resizebox{0.6\columnwidth}{!}{\begin{tabular}{c|c|c|c|c|c}
\hline
\textbf{Defense} & $\alpha_{train}$ &$\alpha_{test}$&  $\mathcal{A}_{\text{rob}}$ (\%) & $\mathcal{A}_{\text{sa}}$ (\%) &$\mathcal{A}_{\text{tr}}$(\%)  \\
\hline
\hline
 PGD & - & 1.0 & 49.29 & 25.09 & 22.91 \\
 PGD+ours  & 2.0 & 1.0 &49.53  & \textbf{26.49} & \textbf{23.94} \\
 \hline
 PGD & - & 1.5 & 49.29 & 17.33 & 15.10\\
 PGD+ours & 2.0 & 1.5 & 49.53 & \textbf{18.92} & \textbf{16.25} \\
 \hline
 PGD& - & 2.0 & 49.29 & 11.66 & 9.72 \\
 PGD+ours & 2.0 & 2.0 & 49.53 & \textbf{13.19} & \textbf{10.81}\\
 \hline
\hline
 TRADES & - & 1.0 & 53.38 & 33.36 & 31.10 \\
 TRADES+ours  & 2.0 & 1.0 &54.10  & \textbf{36.36} & \textbf{33.26}\\
 \hline
 TRADES & - & 1.5 &  53.38 & 25.92 & 23.31\\
 TRADES+ours & 2.0 & 1.5 & 54.10 & \textbf{29.52} & \textbf{25.84}\\
 \hline
 TRADES & - & 2.0 &  53.38 & 19.78 & 17.14 \\
 TRADES+ours & 2.0 & 2.0 & 54.10 & \textbf{23.62} & \textbf{19.79}\\
 \hline
\end{tabular}}
\end{table}
\begin{table}[!htbp]
\centering
\caption{\small \textbf{Robustness against non-uniform attacks on Tiny ImageNet (Proposed Metrics).} The adversarial examples are generated through 10-PGD with $\epsilon=0.016$.}  
\label{tab:adv_eps_compare_trades_tr_tinyimagenet}
\vspace{-0.5em}
\resizebox{0.6\columnwidth}{!}{\begin{tabular}{c|c|c|c|c|c}
\hline
\textbf{Defense} & $\alpha_{train}$ &$\alpha_{test}$&  $\mathcal{A}_{\text{rob}}$ (\%) &  $\mathcal{A}_{\text{sa}}$ (\%)  &$\mathcal{A}_{\text{tr}}$(\%)  \\
\hline
\hline
 PGD  & - & 0.5 & 22.27 & 17.91 & 17.14 \\
 PGD+ours  & 0.3 & 0.5 & 22.75 & \textbf{19.25} & \textbf{18.53}\\
\hline
 PGD  & - & 1.0 & 22.27 & 14.33 & 12.98 \\
 PGD+ours  & 0.3 & 1.0 & 22.75 & \textbf{16.34} & \textbf{14.99}\\
 \hline
 PGD  & - & 1.5 & 22.27 & 11.42 & 9.79 \\
 PGD+ours  & 0.3 & 1.5 & 22.75 & \textbf{13.88} & \textbf{12.06}\\
 \hline
 PGD  & - & 2.0 & 22.27 & 9.04 & 7.32 \\
 PGD+ours  & 0.3 & 2.0 & 22.75 & \textbf{11.80} & \textbf{9.66}\\
\hline
\hline
 TRADES  & - & 0.5 & 27.90 & 23.95 & 22.94 \\
 TRADES+ours  & 5.0 & 0.5 & 28.16 & \textbf{25.18} & \textbf{24.25} \\
 \hline
 TRADES  & - & 1.0 & 27.90 & 20.57 & 18.87 \\
 TRADES+ours  & 5.0 & 1.0 & 28.16 & \textbf{22.40} & \textbf{20.74} \\
 \hline
 TRADES & - & 1.5 & 27.90 & 17.74 &15.33 \\
 TRADES+ours  & 5.0 & 1.5 & 28.16 & \textbf{20.02} &\textbf{17.71} \\
 \hline
 TRADES  & - & 2.0 & 27.90 & 15.33 & 12.58\\
 TRADES+ours  & 5.0 & 2.0 & 28.16 & \textbf{17.79} & \textbf{15.01}\\
\hline
\end{tabular}}
\end{table}

\subsection{Experimental Results and Analysis}
In this section, we firstly show that while the robust network trained using unweighted adversarial training objective will fail in the presence of non-uniform attacks, the network trained by our defense mechanism is able to withstand the strong non-uniform attacks. Then, we verify that our proposed re-weighting approach, although designed for stronger non-uniform attacks, matches the state-of-the-art adversarial training based algorithms even in traditional uniform attack settings.
Experiments are conducted on MNIST \citep{lecun1998mnist}, CIFAR10 \citep{Krizhevsky2012cifar10} and Tiny ImageNet \citep{le2015tiny} datasets. 
Finally, we evaluate the trained models under different attack algorithms and the DRO setting \citep{staib2017distributionally}. The detailed results of these experiments are provided in Appendix~\ref{appendix:different_attacks}, Appendix~\ref{appendix:change_alpha} and Appendix~\ref{appendix:viz}.

\noindent \textbf{Baselines and experimental settings} We use adversarial training \citep{madry2017towards} and TRADES \citep{zhang2019theoretically} as baselines. In the context of TRADES, the robust regularization term is governed by a penalty strength $\lambda$. Moreover, regarding CIFAR10, we also include our reproduced results of IAAT~\citep{balaji2019instance}, YOPO~\citep{zhang2019you} and AT4Free~\citep{shafahi2019adversarial}. We then conduct ablation studies (results in Appendix~\ref{appendix:sensi_hyper}) of the re-weighting approaches on top of the loss function of the baselines. The detailed experiment settings are described in Appendix~\ref{appendix:exp-setting}.

\noindent \textbf{Robustness under non-uniform attack} As argued previously, the core of this work is that the minimax optimization objective for adversarial loss should take the distribution of adversarial examples into account and it should help the network defend against non-uniform attackers. Now, we show that the models trained with traditional adversarial training algorithms (PGD-based adversarial training and TRADES) will perform poorly in the presence of a non-uniform attacker whereas our method is able to better defend against such non-uniform attacker. The experimental results are demonstrated in Table~\ref{tab:adv_eps_compare_trades_tr_mnist}, Table~\ref{tab:adv_eps_compare_trades_tr_cifar10} and Table~\ref{tab:adv_eps_compare_trades_tr_tinyimagenet}. For a given $\alpha_{attack}$, we observe that $\mathcal{A}_{sa}$ and $\mathcal{A}_{tr}$ are smaller than $\mathcal{A}_{robust}$ in most cases on all datasets. For instance, on MNIST, if the attacker scales the margin error by $2$, i.e. $\alpha_{attack} =2.0$, and it generates and sample adversarial examples using the rescaled margin error, the accuracy on adversarial examples of baseline TRADES model will decrease dramatically, with $\mathcal{A}_{rob} = 95.59\%$ dropping to $\mathcal{A}_{sa} =52.15\%$ and $\mathcal{A}_{tr} = 51.52\%$.

Moreover, for a trained model (trained with a specific $\alpha_{train}$), if the $\alpha_{test}$ goes larger, indicating that the attacker is more powerful, $\mathcal{A}_{sa}$ and $\mathcal{A}_{tr}$ will drop even further. However, faced with the same non-uniform attacker, our model is able to achieve better robustness. For example, on MNIST, if trained with $\alpha_{train}=2.0$, the modified TRADES model is able to achieve $\mathcal{A}_{sa} = 66.71\%$ and $\mathcal{A}_{tr} = 65.61\%$ when defending against $\alpha_{attack}=2.0$, in comparison to $\mathcal{A}_{sa} =52.15\%$ and $\mathcal{A}_{tr} = 51.52\%$ of the baseline method. Our model is able to consistently beat the baselines under varying $\alpha_{attack}$'s for all tested datasets.

\begin{table}[!htbp]
\centering
\caption{\small \textbf{Natural error and robust error under uniform attacks on CIFAR10 (Traditional Metrics).} The adversarial examples are generated through 20-PGD with $\epsilon=0.031$.} 
\label{tab:adv_eps_compare_trades_cifar10}
\vspace{-0.5em}
\resizebox{0.55\columnwidth}{!}{\begin{tabular}{c|c|c|c|c}
\hline
\textbf{Defense} & $1/\lambda$ & $\alpha_{train}$ & $\mathcal{A}_{\text{nat}}$ (\%)  &$\mathcal{A}_{\text{rob}}$(\%) \\
\hline
 \begin{tabular}{c}
    AT4Free   \\
\end{tabular}  & - & - & 81.80 &39.00\\
 \begin{tabular}{c}
    YOPO-5-3   \\
\end{tabular}  & - & - & 83.99 &44.72\\
 \begin{tabular}{c}
    IAAT  \\
\end{tabular}  & - & - & 88.60 &48.27\\
\hline
\hline
 \begin{tabular}{c}
    PGD   \\
\end{tabular}  & - & - & 82.00 &49.29\\
\hline
 PGD+ours &- &0.01 &82.33 & 49.08\\ 
 PGD+ours &- &0.05 &81.75 & 49.25\\ 
 PGD+ours &- &0.1 &81.60 &49.53 \\
\hline
\hline
  \begin{tabular}{c}
    TRADES   \\
\end{tabular} & 5 & - & 82.93 & 53.38 \\
\hline
 TRADES+ours &5 &0.1 &82.98 &54.10\\ 
 TRADES+ours &5 &1.0 &83.17 &54.05 \\
 TRADES+ours &5 &1.5 &82.83 &53.91 \\
 TRADES+ours &5 &2.0 &83.41 &54.10 \\
\hline

\end{tabular}}
\vspace{-1em}
\end{table}

\noindent \textbf{Robustness under uniform attack} Comparing the results in Table~\ref{tab:adv_eps_compare_trades_minst}, Table~\ref{tab:adv_eps_compare_trades_cifar10} and Table~\ref{tab:adv_eps_compare_trades_tinyimagenet}, our modified defense mechanism,designed for non-uniform attacks, matches or slightly outperforms the state-of-the-art uniform attacks.  
On CIFAR10, the best robust accuracy of TRADES-trained model using our method is 54.10\%, which is better than 53.38\% of the baseline model. Actually, we are able to obtain similar observations from the results on MNIST and Tiny ImageNet on models trained using PGD and TRADES. To summarize, uniform attack results show that our modified training objective maintain the performance under uniform attacks and might even increase the performance of the trained models under traditional metrics. 

\begin{table}[!htbp]
\centering
\caption{\small \textbf{Natural error and robust error under uniform attacks on Tiny ImageNet (Traditional Metrics).} The adversarial examples are generated via 10-PGD with $\epsilon=0.016$.} 
\label{tab:adv_eps_compare_trades_tinyimagenet}
\vspace{-0.5em}
\resizebox{0.55\columnwidth}{!}{\begin{tabular}{c|c|c|c|c}
\hline
\textbf{Defense} & $1/\lambda$ & $\alpha_{train}$ & $\mathcal{A}_{\text{nat}}$ (\%)  &$\mathcal{A}_{\text{rob}}$(\%) \\
\hline
 \begin{tabular}{c}
    PGD   \\
\end{tabular}  & - & - & 35.02 &22.27\\
 \hline
 PGD+ours &6 &0.1 &35.76 &23.16 \\ 
 PGD+ours &6 &0.15 &34.60 &22.17 \\ 
 PGD+ours &6 &0.2 &35.52 &22.75 \\
 PGD+ours &6 &0.25 &33.10 &21.36 \\
 PGD+ours &6 &0.3 &34.26 &22.75 \\
 \hline
 \hline
 \begin{tabular}{c}
   TRADES    \\
\end{tabular}  & 6 & - &45.44  &27.90 \\
 \hline
 TRADES+ours &6 &0.5 &44.58 &28.28 \\ 
 TRADES+ours &6 &1.5 &45.64 &28.24 \\
 TRADES+ours &6 &2.0 &45.72 &28.74 \\
 TRADES+ours &6 &2.5 &45.34 &28.44 \\
 TRADES+ours &6 &3.5 &45.35 &28.26 \\
 TRADES+ours &6 &5.0 &45.15 &28.16 \\
 \hline
\end{tabular}}
\end{table}
\section{Conclusion}\label{sec:conc}
This work studies the objective function for adversarial training. We argue that adversarial examples are not all created equal, and therefore the loss function should learn to weigh the individual examples during training. 
Our method improves the performance of both clean data natural accuracy and robust accuracy of the baseline under both uniform and non-uniform attack schemes. The learnable weighted minimax risk motivates us to analyze the adversarial risk from a different perspective. That is, we should introduce flexibility to the model and let it assign different penalties to the individual data points during adversarial training.



\bibliographystyle{unsrtnat}
\bibliography{references.bib}

\begin{thebibliography}{41}
\providecommand{\natexlab}[1]{#1}
\providecommand{\url}[1]{\texttt{#1}}
\expandafter\ifx\csname urlstyle\endcsname\relax
  \providecommand{\doi}[1]{doi: #1}\else
  \providecommand{\doi}{doi: \begingroup \urlstyle{rm}\Url}\fi

\bibitem[Szegedy et~al.(2013)Szegedy, Zaremba, Sutskever, Bruna, Erhan,
  Goodfellow, and Fergus]{szegedy2013intriguing}
Christian Szegedy, Wojciech Zaremba, Ilya Sutskever, Joan Bruna, Dumitru Erhan,
  Ian Goodfellow, and Rob Fergus.
\newblock Intriguing properties of neural networks.
\newblock \emph{arXiv preprint arXiv:1312.6199}, 2013.

\bibitem[Huang et~al.(2017)Huang, Papernot, Goodfellow, Duan, and
  Abbeel]{huang2017adversarial}
Sandy Huang, Nicolas Papernot, Ian Goodfellow, Yan Duan, and Pieter Abbeel.
\newblock Adversarial attacks on neural network policies.
\newblock \emph{arXiv preprint arXiv:1702.02284}, 2017.

\bibitem[Zhang et~al.(2019{\natexlab{a}})Zhang, Yu, Jiao, Xing, Ghaoui, and
  Jordan]{zhang2019theoretically}
Hongyang Zhang, Yaodong Yu, Jiantao Jiao, Eric~P Xing, Laurent~El Ghaoui, and
  Michael~I Jordan.
\newblock Theoretically principled trade-off between robustness and accuracy.
\newblock \emph{arXiv preprint arXiv:1901.08573}, 2019{\natexlab{a}}.

\bibitem[Carlini and Wagner(2017)]{carlini2017adversarial}
Nicholas Carlini and David Wagner.
\newblock Adversarial examples are not easily detected: Bypassing ten detection
  methods.
\newblock In \emph{Proceedings of the 10th ACM Workshop on Artificial
  Intelligence and Security}, pages 3--14, 2017.

\bibitem[Kannan et~al.(2018)Kannan, Kurakin, and
  Goodfellow]{Kannan2018AdversarialLP}
Harini Kannan, Alexey Kurakin, and Ian~J. Goodfellow.
\newblock Adversarial logit pairing.
\newblock \emph{ArXiv}, abs/1803.06373, 2018.

\bibitem[Kurakin et~al.(2016)Kurakin, Goodfellow, and
  Bengio]{kurakin2016adversarial}
Alexey Kurakin, Ian Goodfellow, and Samy Bengio.
\newblock Adversarial machine learning at scale.
\newblock \emph{arXiv preprint arXiv:1611.01236}, 2016.

\bibitem[Shaham et~al.(2018)Shaham, Yamada, and
  Negahban]{shaham2018understanding}
Uri Shaham, Yutaro Yamada, and Sahand Negahban.
\newblock Understanding adversarial training: Increasing local stability of
  supervised models through robust optimization.
\newblock \emph{Neurocomputing}, 307:\penalty0 195--204, 2018.

\bibitem[Shafahi et~al.(2019{\natexlab{a}})Shafahi, Najibi, Ghiasi, Xu,
  Dickerson, Studer, Davis, Taylor, and Goldstein]{shafahi2019adversarial}
Ali Shafahi, Mahyar Najibi, Mohammad~Amin Ghiasi, Zheng Xu, John Dickerson,
  Christoph Studer, Larry~S Davis, Gavin Taylor, and Tom Goldstein.
\newblock Adversarial training for free!
\newblock In \emph{Advances in Neural Information Processing Systems}, pages
  3353--3364, 2019{\natexlab{a}}.

\bibitem[Zhang et~al.(2019{\natexlab{b}})Zhang, Zhang, Lu, Zhu, and
  Dong]{zhang2019propagate}
Dinghuai Zhang, Tianyuan Zhang, Yiping Lu, Zhanxing Zhu, and Bin Dong.
\newblock You only propagate once: Accelerating adversarial training via
  maximal principle, 2019{\natexlab{b}}.

\bibitem[Staib and Jegelka(2017)]{staib2017distributionally}
Matthew Staib and Stefanie Jegelka.
\newblock Distributionally robust deep learning as a generalization of
  adversarial training.
\newblock In \emph{NIPS workshop on Machine Learning and Computer Security},
  2017.

\bibitem[He et~al.(2016)He, Zhang, Ren, and Sun]{He2016ResNet}
Kaiming He, Xiangyu Zhang, Shaoqing Ren, and Jian Sun.
\newblock Deep residual learning for image recognition.
\newblock In \emph{The IEEE Conference on Computer Vision and Pattern
  Recognition (CVPR)}, June 2016.

\bibitem[Ma et~al.(2018)Ma, Li, Wang, Erfani, Wijewickrema, Schoenebeck, Song,
  Houle, and Bailey]{ma2018characterizing}
Xingjun Ma, Bo~Li, Yisen Wang, Sarah~M Erfani, Sudanthi Wijewickrema, Grant
  Schoenebeck, Dawn Song, Michael~E Houle, and James Bailey.
\newblock Characterizing adversarial subspaces using local intrinsic
  dimensionality.
\newblock \emph{arXiv preprint arXiv:1801.02613}, 2018.

\bibitem[Meng and Chen(2017)]{meng2017magnet}
Dongyu Meng and Hao Chen.
\newblock Magnet: a two-pronged defense against adversarial examples.
\newblock In \emph{Proceedings of the 2017 ACM SIGSAC Conference on Computer
  and Communications Security}, pages 135--147, 2017.

\bibitem[Xu et~al.(2017)Xu, Evans, and Qi]{xu2017feature}
Weilin Xu, David Evans, and Yanjun Qi.
\newblock Feature squeezing: Detecting adversarial examples in deep neural
  networks.
\newblock \emph{arXiv preprint arXiv:1704.01155}, 2017.

\bibitem[Mosbach et~al.(2018)Mosbach, Andriushchenko, Trost, Hein, and
  Klakow]{mosbach2018logit}
Marius Mosbach, Maksym Andriushchenko, Thomas Trost, Matthias Hein, and
  Dietrich Klakow.
\newblock Logit pairing methods can fool gradient-based attacks.
\newblock \emph{arXiv preprint arXiv:1810.12042}, 2018.

\bibitem[Shafahi et~al.(2019{\natexlab{b}})Shafahi, Ghiasi, Huang, and
  Goldstein]{shafahi2019label}
Ali Shafahi, Amin Ghiasi, Furong Huang, and Tom Goldstein.
\newblock Label smoothing and logit squeezing: A replacement for adversarial
  training?
\newblock \emph{arXiv preprint arXiv:1910.11585}, 2019{\natexlab{b}}.

\bibitem[Elsayed et~al.(2018)Elsayed, Krishnan, Mobahi, Regan, and
  Bengio]{elsayed2018large}
Gamaleldin Elsayed, Dilip Krishnan, Hossein Mobahi, Kevin Regan, and Samy
  Bengio.
\newblock Large margin deep networks for classification.
\newblock In \emph{Advances in neural information processing systems}, pages
  842--852, 2018.

\bibitem[Finlay and Oberman(2019)]{finlay2019scaleable}
Chris Finlay and Adam~M Oberman.
\newblock Scaleable input gradient regularization for adversarial robustness.
\newblock \emph{arXiv preprint arXiv:1905.11468}, 2019.

\bibitem[Ross and Doshi-Velez(2018)]{ross2018improving}
Andrew~Slavin Ross and Finale Doshi-Velez.
\newblock Improving the adversarial robustness and interpretability of deep
  neural networks by regularizing their input gradients.
\newblock In \emph{Thirty-second AAAI conference on artificial intelligence},
  2018.

\bibitem[Qin et~al.(2019)Qin, Martens, Gowal, Krishnan, Dvijotham, Fawzi, De,
  Stanforth, and Kohli]{qin2019adversarial}
Chongli Qin, James Martens, Sven Gowal, Dilip Krishnan, Krishnamurthy
  Dvijotham, Alhussein Fawzi, Soham De, Robert Stanforth, and Pushmeet Kohli.
\newblock Adversarial robustness through local linearization.
\newblock In \emph{Advances in Neural Information Processing Systems}, pages
  13824--13833, 2019.

\bibitem[Jakubovitz and Giryes(2018)]{jakubovitz2018improving}
Daniel Jakubovitz and Raja Giryes.
\newblock Improving dnn robustness to adversarial attacks using jacobian
  regularization.
\newblock In \emph{Proceedings of the European Conference on Computer Vision
  (ECCV)}, pages 514--529, 2018.

\bibitem[Madry et~al.(2017)Madry, Makelov, Schmidt, Tsipras, and
  Vladu]{madry2017towards}
Aleksander Madry, Aleksandar Makelov, Ludwig Schmidt, Dimitris Tsipras, and
  Adrian Vladu.
\newblock Towards deep learning models resistant to adversarial attacks.
\newblock \emph{arXiv preprint arXiv:1706.06083}, 2017.

\bibitem[Athalye et~al.(2018)Athalye, Carlini, and
  Wagner]{athalye2018obfuscated}
Anish Athalye, Nicholas Carlini, and David Wagner.
\newblock Obfuscated gradients give a false sense of security: Circumventing
  defenses to adversarial examples, 2018.

\bibitem[Balaji et~al.(2019)Balaji, Goldstein, and Hoffman]{balaji2019instance}
Yogesh Balaji, Tom Goldstein, and Judy Hoffman.
\newblock Instance adaptive adversarial training: Improved accuracy tradeoffs
  in neural nets.
\newblock \emph{arXiv preprint arXiv:1910.08051}, 2019.

\bibitem[Zhu et~al.(2020)Zhu, Cheng, Gan, Sun, Goldstein, and
  Liu]{zhu2019freelb}
Chen Zhu, Yu~Cheng, Zhe Gan, Siqi Sun, Tom Goldstein, and Jingjing Liu.
\newblock Freelb: Enhanced adversarial training for natural language
  understanding.
\newblock In \emph{International Conference on Learning Representations}, 2020.

\bibitem[Jiang et~al.(2020)Jiang, He, Chen, Liu, Gao, and Zhao]{jiang2019smart}
Haoming Jiang, Pengcheng He, Weizhu Chen, Xiaodong Liu, Jianfeng Gao, and Tuo
  Zhao.
\newblock Smart: Robust and efficient fine-tuning for pre-trained natural
  language models through principled regularized optimization.
\newblock \emph{ACL}, 2020.

\bibitem[Gan et~al.(2020)Gan, Chen, Li, Zhu, Cheng, and Liu]{gan2020large}
Zhe Gan, Yen-Chun Chen, Linjie Li, Chen Zhu, Yu~Cheng, and Jingjing Liu.
\newblock Large-scale adversarial training for vision-and-language
  representation learning.
\newblock \emph{NeurIPS}, 2020.

\bibitem[Kong et~al.(2020)Kong, Li, Ding, Wu, Zhu, Ghanem, Taylor, and
  Goldstein]{kong2020flag}
Kezhi Kong, Guohao Li, Mucong Ding, Zuxuan Wu, Chen Zhu, Bernard Ghanem, Gavin
  Taylor, and Tom Goldstein.
\newblock Flag: Adversarial data augmentation for graph neural networks.
\newblock \emph{arXiv preprint arXiv:2010.09891}, 2020.

\bibitem[Ben-Tal et~al.(2013)Ben-Tal, Den~Hertog, De~Waegenaere, Melenberg, and
  Rennen]{ben2013robust}
Aharon Ben-Tal, Dick Den~Hertog, Anja De~Waegenaere, Bertrand Melenberg, and
  Gijs Rennen.
\newblock Robust solutions of optimization problems affected by uncertain
  probabilities.
\newblock \emph{Management Science}, 59\penalty0 (2):\penalty0 341--357, 2013.

\bibitem[Blanchet et~al.(2017)Blanchet, Kang, Zhang, He, and
  Hu]{blanchet2017doubly}
Jose Blanchet, Yang Kang, Fan Zhang, Fei He, and Zhangyi Hu.
\newblock Doubly robust data-driven distributionally robust optimization.
\newblock \emph{arXiv preprint arXiv:1705.07168}, 2017.

\bibitem[Delage and Ye(2010)]{delage2010distributionally}
Erick Delage and Yinyu Ye.
\newblock Distributionally robust optimization under moment uncertainty with
  application to data-driven problems.
\newblock \emph{Operations research}, 58\penalty0 (3):\penalty0 595--612, 2010.

\bibitem[Duchi et~al.(2016)Duchi, Glynn, and Namkoong]{duchi2016statistics}
John Duchi, Peter Glynn, and Hongseok Namkoong.
\newblock Statistics of robust optimization: A generalized empirical likelihood
  approach.
\newblock \emph{arXiv preprint arXiv:1610.03425}, 2016.

\bibitem[Gao and Kleywegt(2016)]{gao2016distributionally}
Rui Gao and Anton~J Kleywegt.
\newblock Distributionally robust stochastic optimization with wasserstein
  distance.
\newblock \emph{arXiv preprint arXiv:1604.02199}, 2016.

\bibitem[Goh and Sim(2010)]{goh2010distributionally}
Joel Goh and Melvyn Sim.
\newblock Distributionally robust optimization and its tractable
  approximations.
\newblock \emph{Operations research}, 58\penalty0 (4-part-1):\penalty0
  902--917, 2010.

\bibitem[Zhang and Liang(2019)]{zhang2019defending}
Yuchen Zhang and Percy Liang.
\newblock Defending against whitebox adversarial attacks via randomized
  discretization.
\newblock \emph{arXiv preprint arXiv:1903.10586}, 2019.

\bibitem[Goodfellow et~al.(2014)Goodfellow, Shlens, and
  Szegedy]{goodfellow2014explaining}
Ian~J Goodfellow, Jonathon Shlens, and Christian Szegedy.
\newblock Explaining and harnessing adversarial examples.
\newblock \emph{arXiv preprint arXiv:1412.6572}, 2014.

\bibitem[Namkoong and Duchi(2016)]{namkoong2016stochastic}
Hongseok Namkoong and John~C Duchi.
\newblock Stochastic gradient methods for distributionally robust optimization
  with f-divergences.
\newblock In \emph{Advances in neural information processing systems}, pages
  2208--2216, 2016.

\bibitem[LeCun(1998)]{lecun1998mnist}
Yann LeCun.
\newblock The mnist database of handwritten digits.
\newblock \emph{http://yann. lecun. com/exdb/mnist/}, 1998.

\bibitem[Krizhevsky(2012)]{Krizhevsky2012cifar10}
Alex Krizhevsky.
\newblock Learning multiple layers of features from tiny images.
\newblock \emph{University of Toronto}, 2012.

\bibitem[Le and Yang(2015)]{le2015tiny}
Ya~Le and Xuan Yang.
\newblock Tiny imagenet visual recognition challenge.
\newblock \emph{CS 231N}, 7, 2015.

\bibitem[Zhang et~al.(2019{\natexlab{c}})Zhang, Zhang, Lu, Zhu, and
  Dong]{zhang2019you}
Dinghuai Zhang, Tianyuan Zhang, Yiping Lu, Zhanxing Zhu, and Bin Dong.
\newblock You only propagate once: Painless adversarial training using maximal
  principle.
\newblock \emph{arXiv preprint arXiv:1905.00877}, 2019{\natexlab{c}}.

\end{thebibliography}

\renewcommand{\appendix}{%
  \par
  \setcounter{section}{0}%
  \renewcommand{\thesection}{A.\arabic{section}}%
}

\clearpage
\appendix
\section{Minimizing Weighted Risk with Mini Batches}\label{appendix:mini}
Regarding the empirical estimation of the expectation of the \emph{adversarial} loss 
\begin{equation}\label{eq:adversarial-training-empirical-one}
    \min_{{\bm{\theta}}}\frac{1}{N}
    \sum\limits_{i=1}^{N}
    \max_{\bm{\delta_{i}}: \| \bm{\delta_{i}} \|<\epsilon} l(f_{\bm{\theta}}(\bm{x_{i}}+\bm{\delta_{i}}), y_{i}),
\end{equation}
people usually use empirical algorithms to derive the \textbf{argmax}. In our work, we use weighted PGD to generate adversarial examples.
\begin{equation}\label{eq:adversarial-training-empirical-two}
    \min_{{\bm{\theta}}}\frac{1}{N}
    \sum\limits_{i=1}^{N}
    \max_{\bm{\delta_{i}}: \| \bm{\delta_{i}} \|<\epsilon}
    s_i(f_{\bm{\theta}}, (\bm{x}_i+\bm{\delta_{i}}), y_i)l(f_{\bm{\theta}}(\bm{x_{i}}+\bm{\delta_{i}}), y_{i})
\end{equation}
Since this procedure is deterministic, let's denote the weighted attack algorithm as $\bm{g}$. We know that the perturbation $\bm{\delta}_i$ w.r.t. a given data point $(\bm{x}_i, y_i)$ is determined not only by the network $f_{\bm{\theta}}$, but also by the loss function $l$. But in our case, we are using weighted loss functions to compute PGD, which are re-scaled independently and the scaling factor is determined by $(\bm{x}_i, y_i, f_{\bm{\theta}}, l)$. Therefore, we can write the perturbations after \textbf{inner maximization} as 
\begin{align}
    \bm{\delta}_i &= \bm{g}(\bm{x}_i, y_i, f_{\bm{\theta}},l) \\ 
     &= \text{arg}\max_{\bm{\delta_{i}}: \| \bm{\delta_{i}} \|<\epsilon}
    \exp\Big(-\alpha \ \text{margin}(f_{\bm{\theta}},\bm{x}_i+\bm{\delta}_i,y_i)\Big)
    \
    l(f_{\bm{\theta}}(\bm{x_{i}}+\bm{\delta_{i}}), y_{i})
\end{align}
\textbf{Scaling Factor.} After we have all the perturbations after K-PGD, we are able to compute the scaling factors for individual loss functions, which is the result of mapping $(\bm{x}_i, y_i, f_{\bm{\theta}} ,\bm{\delta}_i)$ to a scalar $s_i$.
\begin{equation}
    s_i = s_i(f_{\bm{\theta}},\bm{x}_i', y_i)
\end{equation}
\textbf{Remark.} Firstly, we know that $(\bm{x}_i, y_i)$s are i.i.d. Therefore, even if PGD is computed with weights, $\bm{\delta}_i$s are still i.i.d. Since there is no normalization, the resulted full-batch loss $\widetilde{\mathcal{L}}(\bm{\theta})$ is just sum of individual weighted losses. Finally, if we regard the scaling factors as random variables, they must be i.i.d. as well.
\begin{align}
 \widetilde{\mathcal{L}}(\bm{\theta}) 
    \quad  &  = \quad \frac{1}{N}\sum\limits_{i=1}^N
      s_i(f_{\bm{\theta}}, \bm{x}_i', y_i)
     l(f_{\bm{\theta}}(\bm{x_{i}}'), y_{i}) \label{eq:adversarial-training-ours-two} \\
     \quad  &  \approx \quad \mathbb{E}_{({\bm{x}}, y)\sim \mathcal{D}}\Big[
      {s(f_{\bm{\theta}}, \bm{x}', y)}
     l(f_{\bm{\theta}}(\bm{x_{}}'), y_{})\Big] \label{eq:adversarial-training-ours-compact-clean-two}
 \end{align}

The i.i.d. statements about full batch optimization are still valid for mini batches. For a mini-batch, we randomly sample data points $(\bm{x}_i, y_i)$s from the training set. The randomness lies in the sampling process of the data points. Now, we show the statistical equivalence of these two settings.
\begin{equation}
    \begin{split}
    \mathbb{E}_{(\bm{x}_i, y_i)\sim \mathcal{D}}[\widetilde{L}_m(\bm{\theta})] &= \mathbb{E}_{(\bm{x}_i, y_i)\sim \mathcal{D}} [\frac{1}{m} \sum_{i=1}^m s_i(f_{\bm{\theta}}, \bm{x}_i', y_i) l(f_{\bm{\theta}}(\bm{x}_i'), \bm{y}_i)] \\
     &= \frac{1}{m} \mathbb{E}_{(\bm{x}_i, y_i)\sim \mathcal{D}} [\sum_{i=1}^m s_i(f_{\bm{\theta}}, \bm{x}_i', y_i) l(f_{\bm{\theta}}(\bm{x}_i'), \bm{y}_i)] \\
     &= \frac{1}{m} \sum_{i=1}^m  \mathbb{E}_{(\bm{x}_i, y_i)\sim \mathcal{D}} [s_i(f_{\bm{\theta}}, \bm{x}_i', y_i)l(f_{\bm{\theta}}(\bm{x}_i'), \bm{y}_i) \\
     &= \frac{1}{m} m  \mathbb{E}_{(\bm{x}_i, y_i)\sim \mathcal{D}}[s_i(f_{\bm{\theta}}, \bm{x}_i', y_i) l(f_{\bm{\theta}}(\bm{x}_i'), \bm{y}_i)] \\
     &=\widetilde{\mathcal{L}}(\bm{\theta}) 
    \end{split}
\end{equation}
\section{Experiment setting and Additional Results}\label{appendix:exp-setting}
\paragraph{MNIST}  
We use the same training setup as the original adversarial training and TRADES defense, but modify the algorithm by dropping in the proposed weighting factors. We use the same CNN architecture in TRADES (four convolutional layers + three fully-connected layers). We set perturbation $\epsilon = 0.3$, perturbation step size $\eta = 0.01$, number of iterations $K = 40$ (40-PGD), batch size $m = 128$, and run 100 epochs on the training dataset. Regarding our proposed training objective, we used various $\alpha_{defense}$, which is used during training. As for evaluation, the adversarial examples are generated by the same attacker, namely, 40-PGD.

\paragraph{CIFAR10} 
We set up the experiments on CIFAR10 with same attack hyperparameters in TRADES. Similarly, we build our method on the original adversarial training framework as well as TRADES. The output size of ResNet-18 was changed to 10 (originally 1000 for ImageNet classification). The radius of the norm ball was set to $\epsilon = 0.031$, perturbation step size $\eta = 0.007$, number of iterations $S = 10$ (10-PGD during training), batch size $m = 128$, number of epoch is $100$. All 
attack hyperparameters are the same as the ones in the original TRADES paper. As for evaluation, we performed 20-PGD as well as 100-PGD. (The results of 100-PGD are organized in Appendix separately.) Regarding our proposed training objective, we used various $\alpha_{defense}$.

\paragraph{Tiny ImageNet} 
We conduct further experiments on Tiny ImageNet. We use a specific residual network architecture, ResNet-50. We set perturbation $\epsilon= 0.016$, perturbation step size $\eta=0.0015$, number of iterations $K = 10$, batch size $m = 64$, and run 100 epochs on the training dataset. As for TRADES, the penalty strength $1/\lambda$ is set to be 6.0. As for evaluation, we use exactly the same attack parameters to test the robustness of adversarially trained networks. Regarding our proposed training objective, we used various $\alpha_{defense}$. During evaluation, the adversarial examples are generated with exactly the same configuration of training (10-PGD).

\begin{table}[!htbp]
\centering
\caption{\small \textbf{Natural error and robust error under uniform attacks on MNIST (Traditional Metrics).} The adversarial examples are generated through 40-PGD with $\epsilon=0.3$.
}
\label{tab:adv_eps_compare_trades_minst}
\resizebox{0.55\columnwidth}{!}{\begin{tabular}{c|c|c|c|c}
\hline
\textbf{Defense} & $1/\lambda$ & $\alpha_{train}$ & $\mathcal{A}_{\text{nat}}$ (\%)  &$\mathcal{A}_{\text{rob}}$(\%) \\
\hline
 \begin{tabular}{c}
   PGD \\
\end{tabular}
 &- &- & 99.47 & 93.95 \\
 \hline
 PGD+ours &- &0.01 & 99.43 & 94.94 \\ 
 PGD+ours &- &0.2 & 99.48 & 95.41  \\
 PGD+ours &- &0.5 & 99.48 & 95.22 \\
  \hline
  \hline
\begin{tabular}{c}
   TRADES    \\
\end{tabular}
& 6 & - & 99.19 & 95.59\\ 
 \hline
 TRADES+ours &6 &0.5 &99.11 &94.69  \\ 
 TRADES+ours &6 &1.0 &99.10 &95.44  \\
 TRADES+ours &6 &1.5 &99.05 &95.09  \\
 TRADES+ours &6 &2.0 &98.97 &95.20  \\
  \hline
\end{tabular}
}
\vspace{-1em}
\end{table}

\section{Sensitivity of Hyperparameters}\label{appendix:sensi_hyper}
In this section, we provide further experimental results to show how $\alpha_{train}$ and $\alpha_{test}$ can affect the performance of trained neural networks. The experiments in this section are conducted on CIFAR10.

\begin{figure}[!htbp]
\centering
    \begin{subfigure}[b]{0.33\textwidth}
      \resizebox{\linewidth}{!}{
\begin{tikzpicture}

\definecolor{color0}{rgb}{0.886274509803922,0.290196078431373,0.2}
\definecolor{color1}{rgb}{0.203921568627451,0.541176470588235,0.741176470588235}
\definecolor{color2}{rgb}{0.596078431372549,0.556862745098039,0.835294117647059}

\begin{axis}[
axis background/.style={fill=white!89.8039215686275!black},
axis line style={white},
legend cell align={left},
legend style={fill opacity=0.8, draw opacity=1, text opacity=1, draw=white!80!black, fill=white!89.8039215686275!black},
tick align=outside,
tick pos=left,
title={10-PGD Adversarial Training},
x grid style={white},
xlabel={test alpha},
xmajorgrids,
xmin=-0.1, xmax=2.1,
xtick style={color=white!33.3333333333333!black},
y grid style={white},
ylabel={weighted accuracy},
ymajorgrids,
ymin=0.303470565378666, ymax=0.779748992621899,
ytick style={color=white!33.3333333333333!black}
]
\addplot [semithick, color0]
table {%
0 0.744199991226196
0.5 0.645692944526672
1 0.536264181137085
1.5 0.42616394162178
2 0.325119584798813
};
\addlegendentry{baseline}
\addplot [semithick, color1]
table {%
0 0.758099973201752
0.5 0.665805995464325
1 0.562405288219452
1.5 0.456607431173325
2 0.357103765010834
};
\addlegendentry{alpha\_0.01}
\addplot [semithick, color2]
table {%
0 0.75220000743866
0.5 0.65925407409668
1 0.555736660957336
1.5 0.450346767902374
2 0.351604372262955
};
\addlegendentry{alpha\_0.05}
\addplot [semithick, white!46.6666666666667!black]
table {%
0 0.751399993896484
0.5 0.660969257354736
1 0.560828745365143
1.5 0.458891689777374
2 0.362810909748077
};
\addlegendentry{alpha\_0.1}
\end{axis}

\end{tikzpicture}}
      \caption{$\epsilon_{test}=0.0078$}
	\end{subfigure}
	\begin{subfigure}[b]{0.33\textwidth}
      \resizebox{\linewidth}{!}{
\begin{tikzpicture}

\definecolor{color0}{rgb}{0.886274509803922,0.290196078431373,0.2}
\definecolor{color1}{rgb}{0.203921568627451,0.541176470588235,0.741176470588235}
\definecolor{color2}{rgb}{0.596078431372549,0.556862745098039,0.835294117647059}

\begin{axis}[
axis background/.style={fill=white!89.8039215686275!black},
axis line style={white},
legend cell align={left},
legend style={fill opacity=0.8, draw opacity=1, text opacity=1, draw=white!80!black, fill=white!89.8039215686275!black},
tick align=outside,
tick pos=left,
title={10-PGD Adversarial Training},
x grid style={white},
xlabel={test alpha},
xmajorgrids,
xmin=-0.1, xmax=2.1,
xtick style={color=white!33.3333333333333!black},
y grid style={white},
ylabel={weighted accuracy},
ymajorgrids,
ymin=0.220602789521217, ymax=0.688361784815788,
ytick style={color=white!33.3333333333333!black}
]
\addplot [semithick, color0]
table {%
0 0.662899971008301
0.5 0.551173746585846
1 0.436933040618896
1.5 0.331330835819244
2 0.241864562034607
};
\addlegendentry{baseline}
\addplot [semithick, color1]
table {%
0 0.666299998760223
0.5 0.559759736061096
1 0.450655609369278
1.5 0.348688334226608
2 0.260605603456497
};
\addlegendentry{alpha\_0.01}
\addplot [semithick, color2]
table {%
0 0.665199995040894
0.5 0.558773517608643
1 0.44976994395256
1.5 0.347838640213013
2 0.25973704457283
};
\addlegendentry{alpha\_0.05}
\addplot [semithick, white!46.6666666666667!black]
table {%
0 0.667100012302399
0.5 0.563984930515289
1 0.458387523889542
1.5 0.359037786722183
2 0.272143423557281
};
\addlegendentry{alpha\_0.1}
\end{axis}

\end{tikzpicture}}
      \caption{$\epsilon_{test}=0.0156$}
	\end{subfigure}
	\begin{subfigure}[b]{0.33\textwidth}
      \resizebox{\linewidth}{!}{
\begin{tikzpicture}

\definecolor{color0}{rgb}{0.886274509803922,0.290196078431373,0.2}
\definecolor{color1}{rgb}{0.203921568627451,0.541176470588235,0.741176470588235}
\definecolor{color2}{rgb}{0.596078431372549,0.556862745098039,0.835294117647059}

\begin{axis}[
axis background/.style={fill=white!89.8039215686275!black},
axis line style={white},
legend cell align={left},
legend style={fill opacity=0.8, draw opacity=1, text opacity=1, draw=white!80!black, fill=white!89.8039215686275!black},
tick align=outside,
tick pos=left,
title={10-PGD Adversarial Training},
x grid style={white},
xlabel={test alpha},
xmajorgrids,
xmin=-0.1, xmax=2.1,
xtick style={color=white!33.3333333333333!black},
y grid style={white},
ylabel={weighted accuracy},
ymajorgrids,
ymin=0.0986765924841166, ymax=0.486948735639453,
ytick style={color=white!33.3333333333333!black}
]
\addplot [semithick, color0]
table {%
0 0.467599987983704
0.5 0.35023906826973
1 0.250498443841934
1.5 0.172908365726471
2 0.116325326263905
};
\addlegendentry{baseline}
\addplot [semithick, color1]
table {%
0 0.462900012731552
0.5 0.349323928356171
1 0.252509862184525
1.5 0.176534324884415
2 0.120429985225201
};
\addlegendentry{alpha\_0.01}
\addplot [semithick, color2]
table {%
0 0.467400014400482
0.5 0.354875981807709
1 0.258399307727814
1.5 0.182080164551735
2 0.125192955136299
};
\addlegendentry{alpha\_0.05}
\addplot [semithick, white!46.6666666666667!black]
table {%
0 0.469300001859665
0.5 0.359835624694824
1 0.265247881412506
1.5 0.189457729458809
2 0.132065817713737
};
\addlegendentry{alpha\_0.1}
\end{axis}

\end{tikzpicture}}
      \caption{$\epsilon_{test}=0.031$}
	\end{subfigure}
	\begin{subfigure}[b]{0.33\textwidth}
      \resizebox{\linewidth}{!}{
\begin{tikzpicture}

\definecolor{color0}{rgb}{0.886274509803922,0.290196078431373,0.2}
\definecolor{color1}{rgb}{0.203921568627451,0.541176470588235,0.741176470588235}
\definecolor{color2}{rgb}{0.596078431372549,0.556862745098039,0.835294117647059}

\begin{axis}[
axis background/.style={fill=white!89.8039215686275!black},
axis line style={white},
legend cell align={left},
legend style={fill opacity=0.8, draw opacity=1, text opacity=1, draw=white!80!black, fill=white!89.8039215686275!black},
tick align=outside,
tick pos=left,
title={TRADES},
x grid style={white},
xlabel={test alpha},
xmajorgrids,
xmin=-0.1, xmax=2.1,
xtick style={color=white!33.3333333333333!black},
y grid style={white},
ylabel={weighted accuracy},
ymajorgrids,
ymin=0.435063825547695, ymax=0.787730319797993,
ytick style={color=white!33.3333333333333!black}
]
\addplot [semithick, color0]
table {%
0 0.767599999904633
0.5 0.696325778961182
1 0.617150008678436
1.5 0.533977627754211
2 0.451094120740891
};
\addlegendentry{baseline}
\addplot [semithick, color1]
table {%
0 0.767000019550323
0.5 0.696720242500305
1 0.619444310665131
1.5 0.538915395736694
2 0.458980143070221
};
\addlegendentry{alpha\_train: 1.0}
\addplot [semithick, color2]
table {%
0 0.767700016498566
0.5 0.700347900390625
1 0.626893758773804
1.5 0.550579428672791
2 0.474606364965439
};
\addlegendentry{alpha\_train: 1.5}
\addplot [semithick, white!46.6666666666667!black]
table {%
0 0.771700024604797
0.5 0.706681370735168
1 0.63601678609848
1.5 0.562655568122864
2 0.489439606666565
};
\addlegendentry{alpha\_train: 2.0}
\end{axis}

\end{tikzpicture}}
      \caption{$\epsilon_{test}=0.0078$}
	\end{subfigure}
	\begin{subfigure}[b]{0.33\textwidth}
      \resizebox{\linewidth}{!}{
\begin{tikzpicture}

\definecolor{color0}{rgb}{0.886274509803922,0.290196078431373,0.2}
\definecolor{color1}{rgb}{0.203921568627451,0.541176470588235,0.741176470588235}
\definecolor{color2}{rgb}{0.596078431372549,0.556862745098039,0.835294117647059}

\begin{axis}[
axis background/.style={fill=white!89.8039215686275!black},
axis line style={white},
legend cell align={left},
legend style={fill opacity=0.8, draw opacity=1, text opacity=1, draw=white!80!black, fill=white!89.8039215686275!black},
tick align=outside,
tick pos=left,
title={TRADES},
x grid style={white},
xlabel={test alpha},
xmajorgrids,
xmin=-0.1, xmax=2.1,
xtick style={color=white!33.3333333333333!black},
y grid style={white},
ylabel={weighted accuracy},
ymajorgrids,
ymin=0.339525800943375, ymax=0.715174931287766,
ytick style={color=white!33.3333333333333!black}
]
\addplot [semithick, color0]
table {%
0 0.691699981689453
0.5 0.608675360679626
1 0.521930634975433
1.5 0.436479091644287
2 0.356600761413574
};
\addlegendentry{baseline}
\addplot [semithick, color1]
table {%
0 0.694400012493134
0.5 0.613549053668976
1 0.529582321643829
1.5 0.447021007537842
2 0.369576424360275
};
\addlegendentry{alpha\_train: 1.0}
\addplot [semithick, color2]
table {%
0 0.694500029087067
0.5 0.617378115653992
1 0.537766814231873
1.5 0.459459722042084
2 0.385490268468857
};
\addlegendentry{alpha\_train: 1.5}
\addplot [semithick, white!46.6666666666667!black]
table {%
0 0.698099970817566
0.5 0.623412191867828
1 0.546443819999695
1.5 0.470616787672043
2 0.398631751537323
};
\addlegendentry{alpha\_train: 2.0}
\end{axis}

\end{tikzpicture}}
      \caption{$\epsilon_{test}=0.0156$}
	\end{subfigure}
	\begin{subfigure}[b]{0.33\textwidth}
      \resizebox{\linewidth}{!}{
\begin{tikzpicture}

\definecolor{color0}{rgb}{0.886274509803922,0.290196078431373,0.2}
\definecolor{color1}{rgb}{0.203921568627451,0.541176470588235,0.741176470588235}
\definecolor{color2}{rgb}{0.596078431372549,0.556862745098039,0.835294117647059}

\begin{axis}[
axis background/.style={fill=white!89.8039215686275!black},
axis line style={white},
legend cell align={left},
legend style={fill opacity=0.8, draw opacity=1, text opacity=1, draw=white!80!black, fill=white!89.8039215686275!black},
tick align=outside,
tick pos=left,
title={TRADES},
x grid style={white},
xlabel={test alpha},
xmajorgrids,
xmin=-0.1, xmax=2.1,
xtick style={color=white!33.3333333333333!black},
y grid style={white},
ylabel={weighted accuracy},
ymajorgrids,
ymin=0.181909205019474, ymax=0.540290026366711,
ytick style={color=white!33.3333333333333!black}
]
\addplot [semithick, color0]
table {%
0 0.514299988746643
0.5 0.42016789317131
1 0.334083408117294
1.5 0.259694933891296
2 0.198199242353439
};
\addlegendentry{baseline}
\addplot [semithick, color1]
table {%
0 0.522700011730194
0.5 0.431987702846527
1 0.348578751087189
1.5 0.275633871555328
2 0.214256346225739
};
\addlegendentry{alpha\_train: 1.0}
\addplot [semithick, color2]
table {%
0 0.523400008678436
0.5 0.4373439848423
1 0.358100712299347
1.5 0.288192361593246
2 0.228480935096741
};
\addlegendentry{alpha\_train: 1.5}
\addplot [semithick, white!46.6666666666667!black]
table {%
0 0.523999989032745
0.5 0.440442770719528
1 0.363354533910751
1.5 0.294965267181396
2 0.236048832535744
};
\addlegendentry{alpha\_train: 2.0}
\end{axis}

\end{tikzpicture}}
      \caption{$\epsilon_{test}=0.031$}
	\end{subfigure}
\caption{We have 4 trained models using 10-PGD with different $\alpha_{train}$ and 4 trained models using TRADES with different $\alpha_{train}$. The adversarial examples are generated by 10-PGD with $\epsilon_{train}=0.031$. During test time, we test the trained models on adversarial examples generated by PGD-20 with various $\epsilon_{test}$.}
\label{fig:fig}
\end{figure}
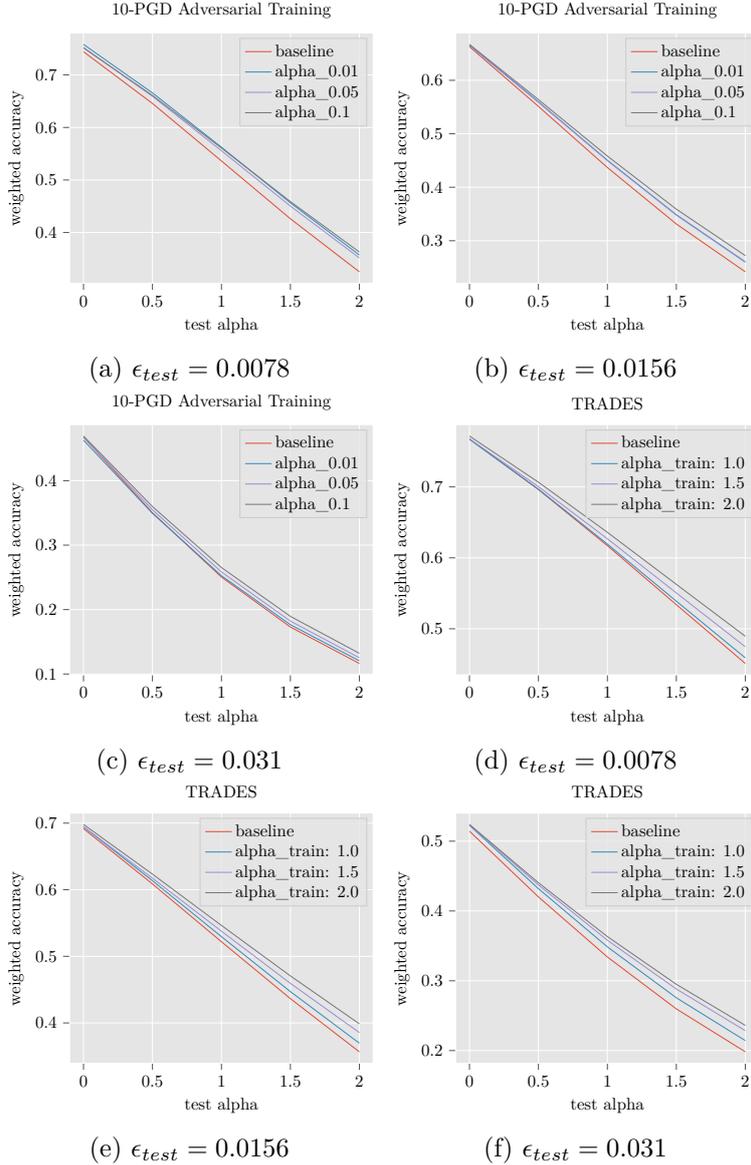

According to Figure~\ref{fig:fig}, the performance of the trained networks drops dramatically in the presence of non-uniform attacks (as $\alpha_{test}$ goes larger), but the model trained with our methods is able to achieve better robustness faced with the same attacker.

\begin{table}[ht]
\centering
\caption{\textbf{PGD Adversarial Training: Robustness against different $\epsilon$ and $\alpha$}. 
The models are trained with a specific $\epsilon_{train}$ and tested with various $\epsilon_{test}$. Like previous setting, $\alpha_{train}$ refers to the $\alpha$ used during training, $\alpha_{test}$ describes the scaling strength of the attacker.} 
\resizebox{0.8\textwidth}{!}{\begin{tabular}{c|c|c|c|c|c|c|c}
\hline
\textbf{Defense} & $\epsilon_{train}$& $\epsilon_{test}$ & $\alpha_{train}$ &$\alpha_{test}$&  $\mathcal{A}_{\text{rob}}$ (\%) & $\mathcal{A}_{\text{sa}}$ (\%) &$\mathcal{A}_{\text{tr}}$(\%)  \\
\hline
 PGD & 0.031 & 0.0078 & - & 1.0 &74.65 &53.67 &52.44 \\
 PGD+ours & 0.031 & 0.0078 & 0.01 & 1.0 &76.06 &56.22 & 54.85\\
 PGD+ours & 0.031 & 0.0078 & 0.05 & 1.0 &76.88 &55.59 & 54.33\\
 PGD+ours & 0.031 & 0.0078 & 0.1 & 1.0 &77.01 &56.08 & 54.74\\
 \hline
 PGD & 0.031 & 0.0078 & - & 1.5 &74.65 &42.66 & 41.02 \\
 PGD+ours & 0.031 & 0.0078 & 0.01 & 1.5 &76.06 &45.64 & 43.90\\
 PGD+ours & 0.031 & 0.0078 & 0.05 & 1.5 &76.88 &45.05 & 43.37\\
 PGD+ours & 0.031 & 0.0078 & 0.1 & 1.5 &77.01 &45.89 & 44.13\\
 \hline
 PGD & 0.031 & 0.0078 & - & 2.0 &74.65 &32.55 & 30.73 \\
 PGD+ours & 0.031 & 0.0078 & 0.01 & 2.0 &76.06 &35.69 & 33.71\\
 PGD+ours & 0.031 & 0.0078 & 0.05 & 2.0 &76.88 &35.17 & 33.26\\
 PGD+ours & 0.031 & 0.0078 & 0.1 & 2.0 &77.01 &36.28 & 34.36\\
\hline
\hline
 PGD & 0.031 & 0.016 & - & 1.0 &66.30 &43.70 & 41.59 \\
 PGD+ours & 0.031 & 0.016 & 0.01 & 1.0 &66.64 &45.08 & 42.97 \\
 PGD+ours & 0.031 & 0.016 & 0.05 & 1.0 &66.51 &44.96 & 43.00 \\
 PGD+ours & 0.031 & 0.016 & 0.1 & 1.0 &66.71 &45.84 & 43.54 \\
\hline
 PGD & 0.031 & 0.016 & - & 1.5 &66.30 &33.14 & 30.67 \\
 PGD+ours & 0.031 & 0.016 & 0.01 & 1.5 &66.64 &34.88 & 32.34 \\
 PGD+ours & 0.031 & 0.016 & 0.05 & 1.5 &66.51 &34.77 & 32.28 \\
 PGD+ours & 0.031 & 0.016 & 0.1 & 1.5 &66.71 &35.90 & 33.10 \\
\hline
 PGD & 0.031 & 0.016 & - & 2.0 &66.30 &24.19 & 21.71 \\
 PGD+ours & 0.031 & 0.016 & 0.01 & 2.0 &66.64 &26.07 & 23.24 \\
 PGD+ours & 0.031 & 0.016 & 0.05 & 2.0 &66.51 &25.96 & 23.36 \\
 PGD+ours & 0.031 & 0.016 & 0.1 & 2.0 &66.71 &27.71 & 23.34 \\
\hline
\hline
 PGD & 0.031 & 0.031 & - & 0.1 &49.29 &44.36 & 44.02 \\
 PGD+ours & 0.031 & 0.031 & 0.01 & 0.1 &49.08 &43.91 & 43.58 \\
 PGD+ours & 0.031 & 0.031 & 0.05 & 0.1 &49.25 &44.41 & 44.01 \\
 PGD+ours & 0.031 & 0.031 & 0.1 & 0.1 &49.53 &44.62 & 44.17 \\
 \hline
 PGD & 0.031 & 0.031 & - & 1.0 &49.29 &25.09 &22.91 \\
 PGD+ours & 0.031 & 0.031 & 0.01 & 1.0 &49.08 &25.24 &23.16 \\
 PGD+ours & 0.031 & 0.031 & 0.05 & 1.0 &49.25 &25.85 &23.62 \\
 PGD+ours & 0.031 & 0.031 & 0.1 & 1.0 &49.53 &26.49 &23.94 \\
  \hline
 PGD & 0.031 & 0.031 & - & 1.5 &49.29 &17.33 &15.10 \\
 PGD+ours & 0.031 & 0.031 & 0.01 & 1.5 &49.08 &17.64 &15.40 \\
 PGD+ours & 0.031 & 0.031 & 0.05 & 1.5 &49.25 &18.22 &15.80 \\
 PGD+ours & 0.031 & 0.031 & 0.1 & 1.5 &49.53 &18.92 &16.25 \\
\hline
\end{tabular}}
\end{table}

\begin{table}[!htbp]
\centering
\caption{\textbf{TRADES: Robustness against different $\epsilon$ and $\alpha$}. 
The models are trained with a specific $\epsilon_{train}$ and tested with various $\epsilon_{test}$. Like previous setting, $\alpha_{train}$ refers to the $\alpha$ used during training, $\alpha_{test}$ describes the scaling strength of the attacker.} 
\resizebox{0.9\textwidth}{!}{\begin{tabular}{c|c|c|c|c|c|c|c}
\hline
\textbf{Defense} & $\epsilon_{train}$& $\epsilon_{test}$ & $\alpha_{train}$ &$\alpha_{test}$&  $\mathcal{A}_{\text{rob}}$ (\%) & $\mathcal{A}_{\text{sa}}$ (\%) &$\mathcal{A}_{\text{tr}}$(\%)  \\
\hline
\hline
 TRADES & 0.031 & 0.0078 & - & 1.0 &76.89 &61.69 & 61.06\\
 TRADES+ours & 0.031 & 0.0078 & 0.1 & 1.0 &76.94 &61.51 & 60.83\\
 TRADES+ours & 0.031 & 0.0078 & 1.0 & 1.0 &76.88 &61.96 & 61.25\\
 TRADES+ours & 0.031 & 0.0078 & 1.5 & 1.0 &77.01 &62.69 & 61.89\\
 TRADES+ours & 0.031 & 0.0078 & 2.0 & 1.0 &77.40 &63.60 & 62.76\\
\hline
 TRADES & 0.031 & 0.0078 & - & 1.5 &76.89 &53.37 & 52.46\\
 TRADES+ours & 0.031 & 0.0078 & 0.1 & 1.5 &76.94 &53.13 & 52.14\\
 TRADES+ours & 0.031 & 0.0078 & 1.0 & 1.5 &76.88 &53.90 & 52.94\\
 TRADES+ours & 0.031 & 0.0078 & 1.5 & 1.5 &77.01 &55.06 & 53.88\\
 TRADES+ours & 0.031 & 0.0078 & 2.0 & 1.5 &77.40 &56.27 & 54.98\\
\hline
 TRADES & 0.031 & 0.0078 & - & 2.0 &76.89 &45.08 & 43.99\\
 TRADES+ours & 0.031 & 0.0078 & 0.1 & 2.0 &76.94 &44.82 & 43.65\\
 TRADES+ours & 0.031 & 0.0078 & 1.0 & 2.0 &76.88 &45.91 & 44.75\\
 TRADES+ours & 0.031 & 0.0078 & 1.5 & 2.0 &77.01 &47.46 & 46.03\\
 TRADES+ours & 0.031 & 0.0078 & 2.0 & 2.0 &77.40 &48.94 & 47.37\\
\hline
\hline
 TRADES+ours & 0.031 & 0.016 & - & 1.0 & 69.62 &52.19 &51.00 \\
 TRADES+ours & 0.031 & 0.016 & 0.1 & 1.0 & 69.77 &52.28 &50.62 \\
 TRADES+ours & 0.031 & 0.016 & 1.0 & 1.0 & 69.83 &52.97 &51.25 \\
 TRADES+ours & 0.031 & 0.016 & 1.5 & 1.0 & 69.82 &53.76 &52.05 \\
 TRADES+ours & 0.031 & 0.016 & 1.5 & 1.0 & 70.21 &54.67 &52.99 \\
 \hline
 TRADES+ours & 0.031 & 0.016 & - & 1.5 & 69.62 &43.65 &42.03 \\
 TRADES+ours & 0.031 & 0.016 & 0.1 & 1.5 & 69.77 &43.70 &41.59 \\
 TRADES+ours & 0.031 & 0.016 & 1.0 & 1.5 & 69.83 &44.72 &42.49 \\
 TRADES+ours & 0.031 & 0.016 & 1.5 & 1.5 & 69.82 &45.93 &43.62 \\
 TRADES+ours & 0.031 & 0.016 & 1.5 & 1.5 & 70.21 &47.09 &44.92 \\
 \hline
 TRADES+ours & 0.031 & 0.016 & - & 2.0 & 69.62 &35.66 &33.73 \\
 TRADES+ours & 0.031 & 0.016 & 0.1 & 2.0 & 69.77 &35.69 &33.31 \\
 TRADES+ours & 0.031 & 0.016 & 1.0 & 2.0 & 69.83 &36.97 &34.42 \\
 TRADES+ours & 0.031 & 0.016 & 1.5 & 2.0 & 69.82 &38.54 &35.87 \\
 TRADES+ours & 0.031 & 0.016 & 1.5 & 2.0 & 70.21 &38.89 &37.51 \\
 \hline
 \hline
 TRADES & 0.031 & 0.031 & - & 1.0 & 53.38 & 33.36 & 31.10 \\
 TRADES+ours & 0.031 & 0.031 & 0.1 & 1.0 & 54.10 & 34.03 &31.83 \\
 TRADES+ours & 0.031 & 0.031 & 1.0 & 1.0 & 54.05 & 34.90 &32.17 \\
 TRADES+ours & 0.031 & 0.031 & 1.5 & 1.0 & 53.91 & 35.62 &32.85 \\
 TRADES+ours & 0.031 & 0.031 & 2.0 & 1.0 & 54.10 & 36.36 &33.26 \\
 \hline
 TRADES & 0.031 & 0.031 & - & 1.5 & 53.38 & 25.92 & 23.31 \\
 TRADES+ours & 0.031 & 0.031 & 0.1 & 1.5 & 54.10 &26.52 &23.90 \\
 TRADES+ours & 0.031 & 0.031 & 1.0 & 1.5 & 54.05 &27.60 &24.38 \\
 TRADES+ours & 0.031 & 0.031 & 1.5 & 1.5 & 53.91 &28.64 &25.25 \\
 TRADES+ours & 0.031 & 0.031 & 2.0 & 1.5 & 54.10 &29.52 &25.84 \\
 \hline
 TRADES & 0.031 & 0.031 & - & 2.0 & 53.38 & 19.78 & 17.14 \\
 TRADES+ours & 0.031 & 0.031 & 0.1 & 2.0 & 54.10 & 20.29 &17.59 \\
 TRADES+ours & 0.031 & 0.031 & 1.0 & 2.0 & 54.05 & 21.46 &18.19 \\
 TRADES+ours & 0.031 & 0.031 & 1.5 & 2.0 & 53.91 & 22.68 &19.09 \\
 TRADES+ours & 0.031 & 0.031 & 2.0 & 2.0 & 54.10 & 23.62 &19.79 \\
\hline
\end{tabular}}
\end{table}
\section{Robustness under Different Attacking Algorithms }\label{appendix:different_attacks}
In this section, we evaluate the trained models under different attacks. Specifically, in all previous sections, the PGD attack is computed based on cross entropy loss. In this section, we introduce another \textbf{margin loss} to compute PGD, which is denoted as \textbf{PGD-margin}. If not specified, \textbf{PGD} refers to the PGD computed via cross entropy loss. The models are mainly tested with 20-PGD, 100-PGD, 20-PGD-margin and 100-PGD-margin. The experiments in this section are conducted on CIFAR10.

According to Figure~\ref{fig:fig}, the performance of the trained networks drops dramatically in the presence of non-uniform attacks (as $\alpha_{test}$ goes larger), but the model trained with our methods is able to achieve better robustness facing with the same attacker.

\begin{table}[ht]
\centering
\caption{\textbf{Robustness under different attacks.}.
The models are trained with a specific $\epsilon_{train}$ and tested with various $\epsilon_{test}$. Like previous setting, $\alpha_{train}$ refers to the $\alpha$ used during training, $\alpha_{test}$ describes the scaling strength of the attacker.} 
\resizebox{\textwidth}{!}{\begin{tabular}{c|c|c|c|c|c|c|c}
\hline
\textbf{Defense} & \textbf{Attack} & $\epsilon_{train}$& $\epsilon_{test}$ & $\alpha_{train}$ &$\alpha_{test}$&  $\mathcal{A}_{\text{rob}}$ (\%) & $\mathcal{A}_{\text{sa}}$ (\%)  \\
\hline
 PGD &20-PGD & 0.031 & 0.031 & - & 1.0 &49.29 &25.09 \\
 PGD+ours &20-PGD & 0.031 & 0.031 & 0.01 & 1.0 &49.08 &25.24 \\
 PGD+ours &20-PGD & 0.031 & 0.031 & 0.05 & 1.0 &49.25 &25.85 \\
 PGD+ours &20-PGD & 0.031 & 0.031 & 0.1 & 1.0 &49.53 &26.49 \\
\hline
 PGD &100-PGD & 0.031 & 0.031 & - & 1.0 &46.77 &25.06 \\
 PGD+ours &100-PGD & 0.031 & 0.031 & 0.01 & 1.0 &46.26 &25.23 \\
 PGD+ours &100-PGD & 0.031 & 0.031 & 0.05 & 1.0 &46.76 &25.86 \\
 PGD+ours &100-PGD & 0.031 & 0.031 & 0.1 & 1.0 &46.94 &26.54 \\
\hline
 PGD &20-PGD-margin & 0.031 & 0.031 & - & 1.0 &45.63 &17.62 \\
 PGD+ours &20-PGD-margin & 0.031 & 0.031 & 0.01 & 1.0 &45.41 &17.90 \\
 PGD+ours &20-PGD-margin & 0.031 & 0.031 & 0.05 & 1.0 &45.95 &18.43 \\
 PGD+ours &20-PGD-margin & 0.031 & 0.031 & 0.1 & 1.0 &45.62 &18.53 \\
\hline
 PGD &100-PGD-margin & 0.031 & 0.031 & - & 1.0 &45.65 &17.64 \\
 PGD+ours &100-PGD-margin & 0.031 & 0.031 & 0.01 & 1.0 &45.41 & 17.90\\
 PGD+ours &100-PGD-margin & 0.031 & 0.031 & 0.05 & 1.0 &45.97 & 18.45\\
 PGD+ours &100-PGD-margin & 0.031 & 0.031 & 0.1 & 1.0 &45.62 & 18.54\\
\hline

\hline
 TRADES &20-PGD & 0.031 & 0.031 & - & 1.0 &53.38 &33.36 \\
 TRADES+ours &20-PGD & 0.031 & 0.031 & 0.1 & 1.0 &54.10 &34.03 \\
 TRADES+ours &20-PGD & 0.031 & 0.031 & 1.0 & 1.0 &54.05 &34.90 \\
 TRADES+ours &20-PGD & 0.031 & 0.031 & 1.5 & 1.0 &53.91 &35.62 \\
 TRADES+ours &20-PGD & 0.031 & 0.031 & 2.0 & 1.0 &54.10 &36.36 \\
 \hline
 TRADES &100-PGD & 0.031 & 0.031 & - & 1.0 &51.44 &33.42 \\
 TRADES+ours &100-PGD & 0.031 & 0.031 & 0.1 & 1.0 &52.09 &34.00 \\
 TRADES+ours &100-PGD & 0.031 & 0.031 & 1.0 & 1.0 &52.24 &34.83 \\
 TRADES+ours &100-PGD & 0.031 & 0.031 & 1.5 & 1.0 &52.27 &35.74 \\
 TRADES+ours &100-PGD & 0.031 & 0.031 & 2.0 & 1.0 &52.44 &36.38 \\
  \hline
 TRADES &20-PGD-margin & 0.031 & 0.031 & - & 1.0 &49.50 &25.64 \\
 TRADES+ours &20-PGD-margin & 0.031 & 0.031 & 0.1 & 1.0 &50.34 & 26.14\\
 TRADES+ours &20-PGD-margin & 0.031 & 0.031 & 1.0 & 1.0 &49.66 & 25.59\\
 TRADES+ours &20-PGD-margin & 0.031 & 0.031 & 1.5 & 1.0 &48.98 & 25.49\\
 TRADES+ours &20-PGD-mragin & 0.031 & 0.031 & 2.0 & 1.0 &48.95 & 25.70\\
 \hline
 TRADES &100-PGD-margin & 0.031 & 0.031 & - & 1.0 &49.49 &25.63 \\
 TRADES+ours &100-PGD-margin & 0.031 & 0.031 & 0.1 &1.0 &50.31 &26.12 \\
 TRADES+ours &100-PGD-margin & 0.031 & 0.031 & 1.0 & 1.0 &49.65 &25.59 \\
 TRADES+ours &100-PGD-margin & 0.031 & 0.031 & 1.5 & 1.0 &48.98 &25.48 \\
 TRADES+ours &100-PGD-mragin & 0.031 & 0.031 & 2.0 & 1.0 &48.93 &25.68 \\
\hline
\end{tabular}}
\end{table}

\begin{figure}[!htbp]
\centering
    \begin{subfigure}[b]{0.4\textwidth}
      \resizebox{\linewidth}{!}{
\begin{tikzpicture}

\definecolor{color0}{rgb}{0.12156862745098,0.466666666666667,0.705882352941177}
\definecolor{color1}{rgb}{1,0.498039215686275,0.0549019607843137}

\begin{axis}[
legend cell align={left},
legend style={fill opacity=0.8, draw opacity=1, text opacity=1, draw=white!80!black},
tick align=outside,
tick pos=both,
title={Weighted accuracy},
x grid style={white!69.0196078431373!black},
xlabel={\(\displaystyle \rho\)},
xmin=-0.0395, xmax=1.0495,
xtick style={color=black},
xtick={-0.2,0,0.2,0.4,0.6,0.8,1,1.2},
xticklabels={−0.2,0.0,0.2,0.4,0.6,0.8,1.0,1.2},
y grid style={white!69.0196078431373!black},
ylabel={accuracy},
ymin=0.0790728763276853, ymax=0.50842157986002,
ytick style={color=black},
ytick={0.05,0.1,0.15,0.2,0.25,0.3,0.35,0.4,0.45,0.5,0.55},
yticklabels={0.05,0.10,0.15,0.20,0.25,0.30,0.35,0.40,0.45,0.50,0.55}
]
\addplot [semithick, color0]
table {%
0.01 0.482053329861871
0.02 0.461281898982289
0.04 0.431906659713575
0.08 0.390363797967201
0.16 0.331613319459566
0.32 0.248573329522419
0.64 0.148771810339774
1 0.098588726488246
};
\addlegendentry{Baseline adversarial}
\addplot [semithick, color1]
table {%
0.01 0.48890572969946
0.02 0.468735902536279
0.04 0.440211459398731
0.08 0.399871805073082
0.16 0.342822918816337
0.32 0.262206962074649
0.64 0.169828811944824
1 0.12288879163105
};
\addlegendentry{Ours adversarial}
\end{axis}

\end{tikzpicture}}
      \caption{Accuracy under Cross Entropy Based PGD Attack}
	\end{subfigure}
	\begin{subfigure}[b]{0.4\textwidth}
      \resizebox{\linewidth}{!}{
\begin{tikzpicture}

\definecolor{color0}{rgb}{0.12156862745098,0.466666666666667,0.705882352941177}
\definecolor{color1}{rgb}{1,0.498039215686275,0.0549019607843137}

\begin{axis}[
legend cell align={left},
legend style={fill opacity=0.8, draw opacity=1, text opacity=1, at={(0.03,0.97)}, anchor=north west, draw=white!80!black},
tick align=outside,
tick pos=both,
title={Weighted loss},
x grid style={white!69.0196078431373!black},
xlabel={\(\displaystyle \rho\)},
xmin=-0.0395, xmax=1.0495,
xtick style={color=black},
xtick={-0.2,0,0.2,0.4,0.6,0.8,1,1.2},
xticklabels={−0.2,0.0,0.2,0.4,0.6,0.8,1.0,1.2},
y grid style={white!69.0196078431373!black},
ylabel={loss},
ymin=1.35763400111773, ymax=2.87010389974988,
ytick style={color=black},
ytick={1.2,1.4,1.6,1.8,2,2.2,2.4,2.6,2.8,3},
yticklabels={1.2,1.4,1.6,1.8,2.0,2.2,2.4,2.6,2.8,3.0}
]
\addplot [semithick, color0]
table {%
0.01 1.44599342555277
0.02 1.51027579851091
0.04 1.60118480215784
0.08 1.72974954802038
0.16 1.91156755529725
0.32 2.16869704716937
0.64 2.52577025514298
1 2.80135526799387
};
\addlegendentry{Baseline adversarial}
\addplot [semithick, color1]
table {%
0.01 1.42638263287374
0.02 1.48636222821408
0.04 1.57118618543611
0.08 1.69114537609848
0.16 1.86079329055748
0.32 2.10071166956851
0.64 2.42797092317523
1 2.67803053200539
};
\addlegendentry{Ours adversarial}
\end{axis}

\end{tikzpicture}}
      \caption{Loss under Cross Entropy Based PGD Attack}
	\end{subfigure}
	\begin{subfigure}[b]{0.4\textwidth}
      \resizebox{\linewidth}{!}{
\begin{tikzpicture}

\definecolor{color0}{rgb}{0.12156862745098,0.466666666666667,0.705882352941177}
\definecolor{color1}{rgb}{1,0.498039215686275,0.0549019607843137}

\begin{axis}[
legend cell align={left},
legend style={fill opacity=0.8, draw opacity=1, text opacity=1, draw=white!80!black},
tick align=outside,
tick pos=both,
title={Weighted accuracy},
x grid style={white!69.0196078431373!black},
xlabel={\(\displaystyle \rho\)},
xmin=-0.0395, xmax=1.0495,
xtick style={color=black},
xtick={-0.2,0,0.2,0.4,0.6,0.8,1,1.2},
xticklabels={−0.2,0.0,0.2,0.4,0.6,0.8,1.0,1.2},
y grid style={white!69.0196078431373!black},
ylabel={accuracy},
ymin=-0.0193585117536232, ymax=0.406528746826086,
ytick style={color=black},
ytick={-0.05,0,0.05,0.1,0.15,0.2,0.25,0.3,0.35,0.4,0.45},
yticklabels={−0.05,0.00,0.05,0.10,0.15,0.20,0.25,0.30,0.35,0.40,0.45}
]
\addplot [semithick, color0]
table {%
0.01 0.376383925913299
0.02 0.37522804375502
0.04 0.371651417629741
0.08 0.320083597397078
0.16 0.244980296003204
0.32 0.135399058044462
0.64 0.00638675490652357
1 0
};
\addlegendentry{Baseline adversarial}
\addplot [semithick, color1]
table {%
0.01 0.387170235072463
0.02 0.386179091385631
0.04 0.371641278534509
0.08 0.320426531091407
0.16 0.242655882451028
0.32 0.129151119154402
0.64 0.0050859563458566
1 0
};
\addlegendentry{Ours adversarial}
\end{axis}

\end{tikzpicture}}
    \caption{Accuracy under Margin Loss Based PGD Attack}
	\end{subfigure}
	\begin{subfigure}[b]{0.4\textwidth}
      \resizebox{\linewidth}{!}{
\begin{tikzpicture}

\definecolor{color0}{rgb}{0.12156862745098,0.466666666666667,0.705882352941177}
\definecolor{color1}{rgb}{1,0.498039215686275,0.0549019607843137}

\begin{axis}[
legend cell align={left},
legend style={fill opacity=0.8, draw opacity=1, text opacity=1, at={(0.03,0.97)}, anchor=north west, draw=white!80!black},
tick align=outside,
tick pos=both,
title={Weighted loss},
x grid style={white!69.0196078431373!black},
xlabel={\(\displaystyle \rho\)},
xmin=-0.0395, xmax=1.0495,
xtick style={color=black},
xtick={-0.2,0,0.2,0.4,0.6,0.8,1,1.2},
xticklabels={−0.2,0.0,0.2,0.4,0.6,0.8,1.0,1.2},
y grid style={white!69.0196078431373!black},
ylabel={loss},
ymin=-0.416167010402502, ymax=2.16584011711392,
ytick style={color=black}
]
\addplot [semithick, color0]
table {%
0.01 -0.0191145582232568
0.02 -0.013224228774483
0.04 0.0050020653416811
0.08 0.265182656198557
0.16 0.628216384698401
0.32 1.117029783164
0.64 1.70564936656714
1 2.04847615677226
};
\addlegendentry{Baseline adversarial}
\addplot [semithick, color1]
table {%
0.01 -0.298803050060847
0.02 -0.292859790251197
0.04 -0.206384171816862
0.08 0.0842031976387877
0.16 0.474957269083198
0.32 0.968441110074815
0.64 1.52050577209573
1 1.83353657447395
};
\addlegendentry{Ours adversarial}
\end{axis}

\end{tikzpicture}}
    \caption{Loss under Margin Loss Based PGD Attack}
	\end{subfigure}
\caption{We also evaluate the distributional robustness under different distribution budgets $\rho$. Our results are using $\alpha=2$ for training. Our approach consistently achieves lower loss than the baseline.}
\label{fig:fig_dro}
\end{figure}
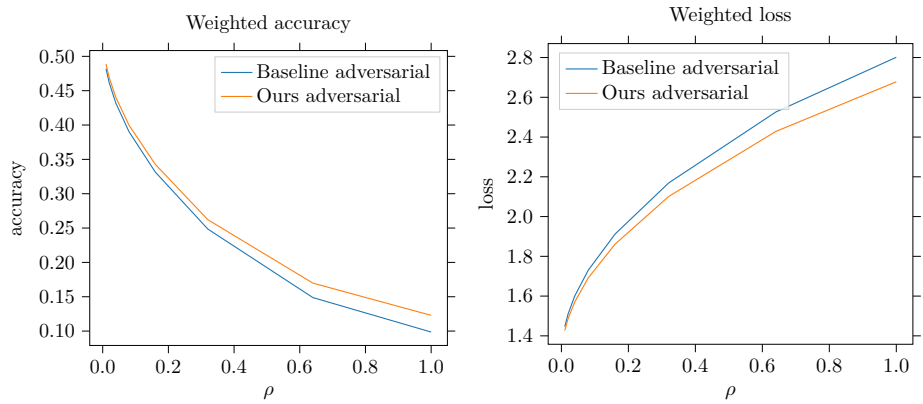
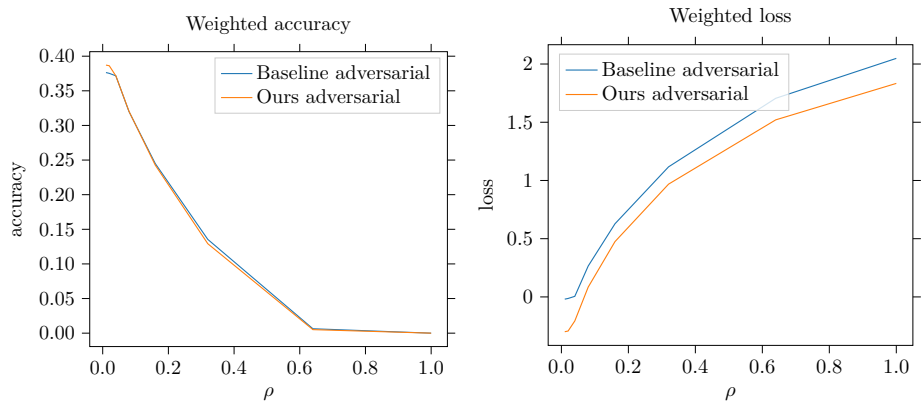

In addition, we compare the robustness of models trained with our weighted loss and the unweighted baseline in the DRO setting, where the adversary has a limited perturbation budget. 
Formally, the adversary tries to solve the following constrained optimization problem
\begin{equation}
\begin{split}
    \max_{\textbf{w}}~& \frac{1}{N} \sum_{i=1}^N w_i l(f(x_i), y_i) \\
    s.t.~ & \sum_{i=1}^N w_i=1, w_i\ge 0, i=1,...,N\\
          & \frac{1}{2} \lVert \textbf{w} - \frac{1}{N} \rVert^2 \le \frac{\rho}{N},
\end{split}
\end{equation}
where $\rho>0$ is the perturbation budget, $w_i$ is the $i$-th entry of the vector $\textbf{w}$, and we are using the $\chi^2$ divergence as the constraint. 
Given the adversarial loss values $l(f(x_i), y_i) $ for each test sample $i$, we solve this constrained optimization problem for different budgets $\rho$, and compute the corresponding weighted adversarial loss and accuracy, as shown in Figure~\ref{fig:fig_dro}. 
Our approach shows consistent improvements over the baseline under all perturbation budgets.

\section{Consistently increasing the robustness.}\label{appendix:change_alpha}
Finally, we show that $\alpha_{defense}$ plays an important roll in terms of increasing the robustness. As shown in Table~\ref{tab:adv_eps_compare_alpha_mnist}, Table~\ref{tab:adv_eps_compare_alpha_cifar10} and Table~\ref{tab:adv_eps_compare_alpha_tinyimagenet}, the consistent increase of $\mathcal{A}_{sa}$ and $\mathcal{A}_{tr}$ with the growth of $\alpha_{defense}$ verifies that margin-aware risk is able to increase the robustness of a network against a uniform attacker and a non-uniform attacker. As mentioned before, normal PGD adversarial training and TRADES are limiting cases of our proposed margin kernel with $\alpha_{defense}=0$. If use a large $\alpha$, the re-weighting effect is enhanced, leading to stronger robustness.

\begin{table}[ht]
\centering
\caption{\textbf{On choosing the weighting strength on MNIST}. 
With this table, we would like to show that re-weighting can indeed increase the robustness of networks. Note that, the adversarial examples are generated via 40-PGD}  
\label{tab:adv_eps_compare_alpha_mnist}
\resizebox{0.65\columnwidth}{!}{\begin{tabular}{c|c|c|c|c|c}
\hline
\textbf{Defense} & $\alpha_{train}$ & $\alpha_{test}$ & $\mathcal{A}_{\text{rob}}$ (\%)  & $\mathcal{A}_{\text{sa}}$ (\%)  &$\mathcal{A}_{\text{tr}}$(\%)  \\

\hline
PGD  &-   &2.0 & 93.95 & 35.32 & 35.04 \\
\hline
PGD+ours  &0.01 &2.0 & 94.94 & \textbf{39.35} & \textbf{38.82}\\
PGD+ours  &0.2 &2.0 & 95.41 & \textbf{43.13} & \textbf{42.62}\\
PGD+ours  &0.1 &2.0 & 95.22 & \textbf{44.96} & \textbf{44.70}\\
\hline
\hline
TRADES  &-   &2.0 & 95.59 & 52.15 & 51.52\\
\hline
TRADES+ours  &0.5 &2.0 & 94.69 & \textbf{53.55} & \textbf{53.01}\\
TRADES+ours  &1.0 &2.0 & 95.44 & \textbf{61.72} & \textbf{61.69}\\
TRADES+ours  &1.5 &2.0 & 95.09 & \textbf{64.75} & \textbf{63.89}\\
TRADES+ours  &2.0 &2.0 & 95.20 & \textbf{66.71} & \textbf{65.61}\\
\hline
\end{tabular}}
\end{table}

\begin{table}[ht]
\centering
\caption{\textbf{On choosing the weighting strength on CIFAR10}. 
With this table, we would like to show that re-weighting can indeed increase the robustness of networks. Note that the adversarial examples are generated via 20-PGD.}  
\label{tab:adv_eps_compare_alpha_cifar10}
\resizebox{0.65\columnwidth}{!}{\begin{tabular}{c|c|c|c|c|c}
\hline
\textbf{Defense} & $\alpha_{train}$ & $\alpha_{test}$ & $\mathcal{A}_{\text{rob}}$ (\%)  & $\mathcal{A}_{\text{sa}}$ (\%)  &$\mathcal{A}_{\text{tr}}$(\%)  \\
\hline
PGD  &-   &2.0 & 49.29 & 11.66 & 9.72\\
\hline
PGD+ours  &0.01 &2.0 & 49.08 & \textbf{12.03} & \textbf{10.01}\\
PGD+ours  &0.05 &2.0 & 49.25 & \textbf{12.53} & \textbf{10.33}\\
PGD+ours  &0.1 &2.0 & 49.53 & \textbf{13.19} & \textbf{10.81}\\
\hline
\hline
TRADES  &-   &2.0 & 53.38 & 19.78 & 17.14\\
\hline
TRADES+ours  &0.1 &2.0 & 54.10 & \textbf{20.29} & \textbf{17.59}\\
TRADES+ours  &1.0 &2.0 & 54.05 & \textbf{21.46} & \textbf{18.19}\\
TRADES+ours  &1.5 &2.0 & 53.91 & \textbf{22.68} & \textbf{19.09}\\
TRADES+ours  &2.0 &2.0 & 54.10 & \textbf{23.62} & \textbf{19.79}\\
\hline
\end{tabular}}
\end{table}

\begin{table}[ht]
\centering
\caption{\textbf{On choosing the weighting strength on Tiny ImageNet}. 
With this table, we would like to show that re-weighting can indeed increase the robustness of networks. Note that the adversarial examples are generated via 10-PGD.}  
\label{tab:adv_eps_compare_alpha_tinyimagenet}
\resizebox{0.65\columnwidth}{!}{\begin{tabular}{c|c|c|c|c|c}
\hline
\textbf{Defense} & $\alpha_{train}$ & $\alpha_{test}$ & $\mathcal{A}_{\text{rob}}$ (\%)  & $\mathcal{A}_{\text{sa}}$ (\%)  &$\mathcal{A}_{\text{tr}}$(\%)  \\
\hline
PGD  &-   &1.0 & 22.27 & 14.33 & 12.98 \\
\hline
PGD+ours  &0.1 &1.0 & 23.16  & \textbf{15.10} & \textbf{13.60}\\
PGD+ours  &0.15 &1.0 & 22.17 & \textbf{15.32} & \textbf{13.88}\\
PGD+ours  &0.2 &1.0 & 22.75 & \textbf{15.69} & \textbf{14.25}\\
PGD+ours  &0.3 &1.0 & 22.75 & \textbf{16.34} & \textbf{14.99}\\
\hline
\hline
TRADES  &-   &1.0 & 27.90 & 20.57 & 18.87 \\
\hline
TRADES+ours  &0.5 &1.0 & 28.28 & \textbf{20.63} & \textbf{19.03}\\
TRADES+ours  &1.5 &1.0 & 28.66 & \textbf{20.86} & \textbf{19.04}\\
TRADES+ours  &2.0 &1.0 & 28.74 & \textbf{21.17} & \textbf{19.50}\\
TRADES+ours  &2.5 &1.0 & 28.44 & \textbf{21.45} & \textbf{19.76}\\
TRADES+ours  &5.0 &1.0 & 28.16 & \textbf{22.40} & \textbf{20.74}\\
\hline
\end{tabular}}
\end{table}
\section{Distribution of The Weights}\label{appendix:viz}
This section, we visualize the distributions of the weights during evaluation under different hyperparameters. The distribution is categorical over all test data points and is formally defined in Definition~\ref{def:scaling}. In each figure, each row correspond to a single TRADES model evaluated on different adversaries with different $\alpha_{test}$. The training setting is specified by the sub-caption of individual sub-figure.

\paragraph{When the attacker gets stronger} According to the plots, for a given trained model, if $\alpha_{test}$ is larger, then the distribution is more spread over the dataset, indicating that the attacker is able to pick data points with larger variety and flexibility. The x-axis refers to the value of the normalized importance score. If the score of a test point is larger, then this point is of greater vulnerability and the attacker is able to sample it more frequently. In an extreme case, where $\alpha_{test}$ is 0, then the distribution is just one column, and the attacker samples all examples uniformly (with an identical probability). When $\alpha_{test}$ is small, we can see that there are not many weights with large values. This observation verifies the intuition that such weaker adversary is ``naive'' and could not perform attacks on vulnerable data points with larger frequency. When $\alpha_{test}$ is large, the attacker is then more powerful, and it tends to utilize the vulnerability of individual samples better. Therefore, we can see there are some weights with large values, meaning that the attacker is able sample vulnerable examples with larger frequency.

\paragraph{When the defense gets stronger} If we observe the plot in columns, the distributions tend to be more concentrated as $\alpha_{train}$ gets larger. For a single column, $\alpha_{test}$ is fixed, but $\alpha_{train}$ varies. The more concentrated distribution means that the attacker fails to attack vulnerable examples with larger frequency. This verifies our statement that the robustness against non-uniform attacker could be obtained using our modified optimization objective. The larger $\alpha_{train}$ is, the more the robustness is enhanced.

\begin{figure}[ht]
\begin{subfigure}{\textwidth}
  \centering
  \includegraphics[width=0.6\linewidth]{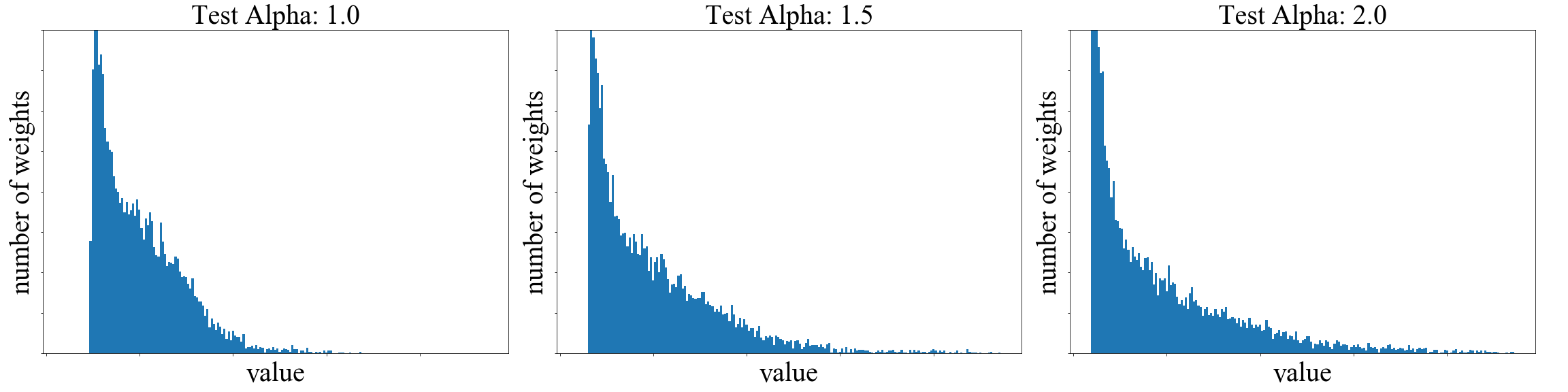}
  \caption{Baseline}
\end{subfigure}
\begin{subfigure}{\textwidth}
  \centering
  \includegraphics[width=0.6\linewidth]{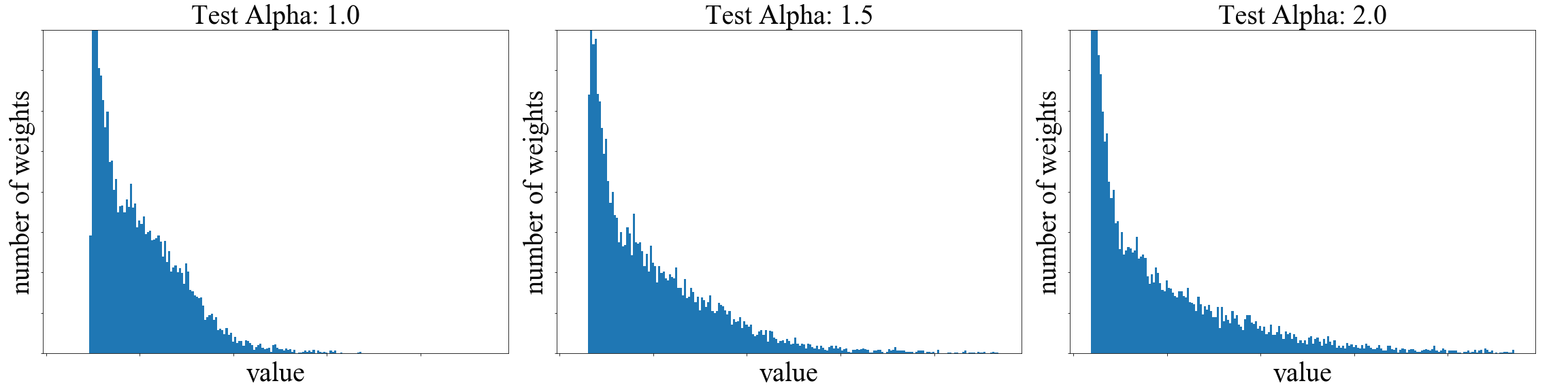}  
  \caption{Training Alpha = 0.1}
\end{subfigure}
\begin{subfigure}{\textwidth}
  \centering
  \includegraphics[width=0.6\linewidth]{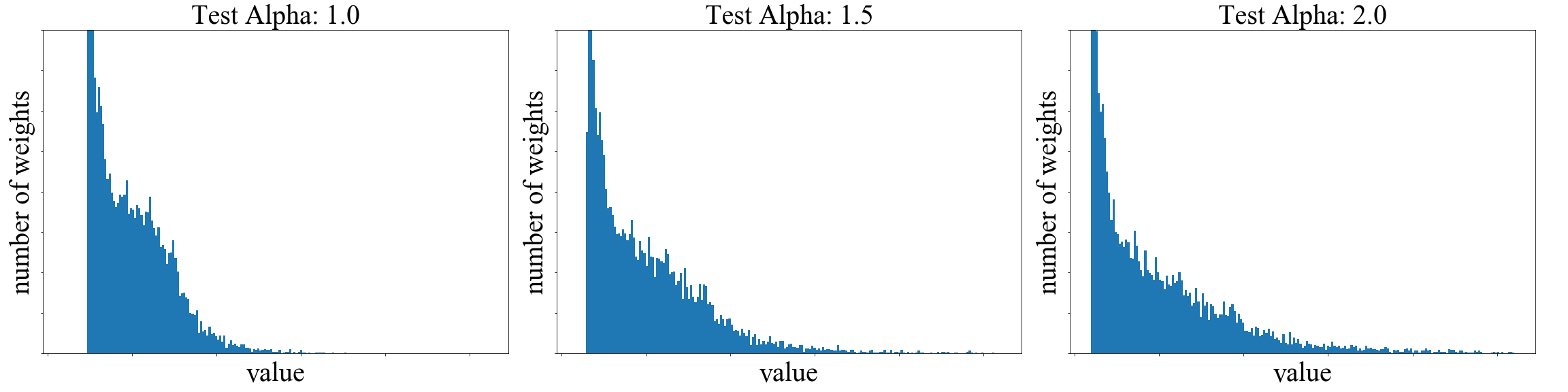}  
  \caption{Training Alpha = 1.0}
\end{subfigure}
\begin{subfigure}{\textwidth}
  \centering
  \includegraphics[width=0.6\linewidth]{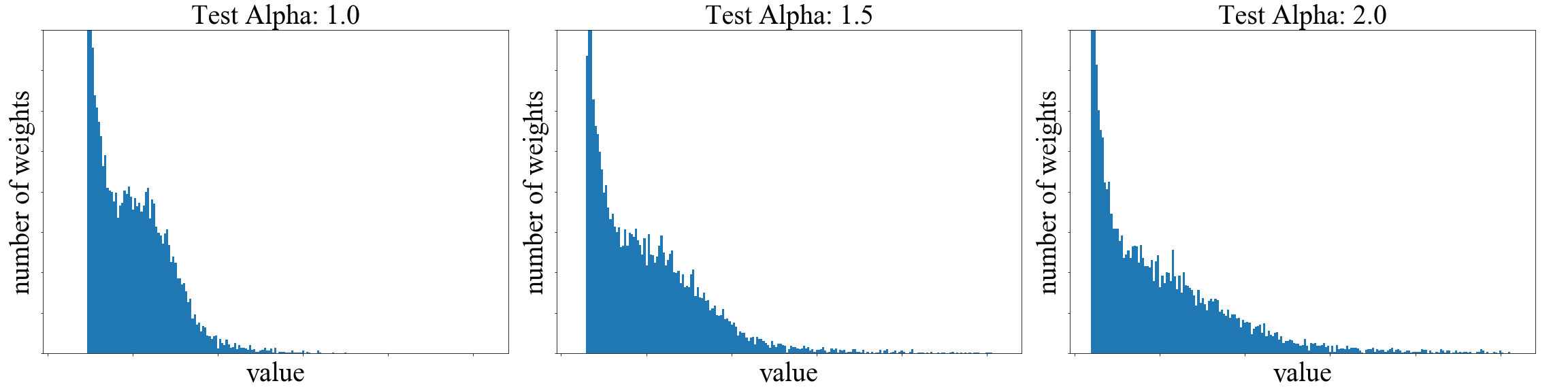}  
  \caption{Training Alpha = 1.5}
\end{subfigure}
\begin{subfigure}{\textwidth}
  \centering
  \includegraphics[width=0.6\linewidth]{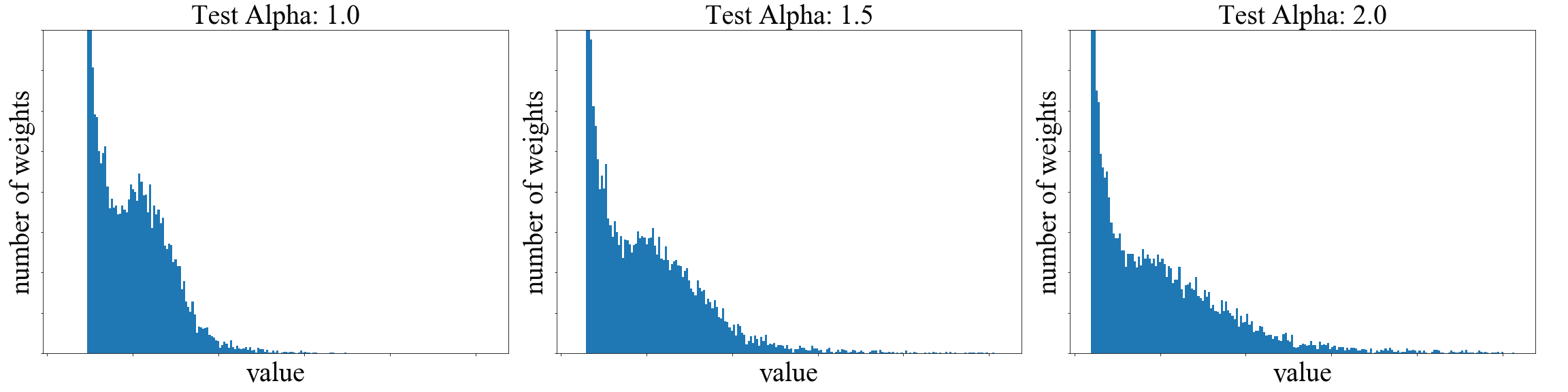}  
  \caption{Training Alpha = 2.0}
\end{subfigure}
\caption{We visualize the importance distribution over the entire test set. The weights are evaluated with $\epsilon=0.0078$}
\label{fig:fig1}
\end{figure}

\begin{figure}[ht]
\begin{subfigure}{\textwidth}
  \centering
  \includegraphics[width=0.6\linewidth]{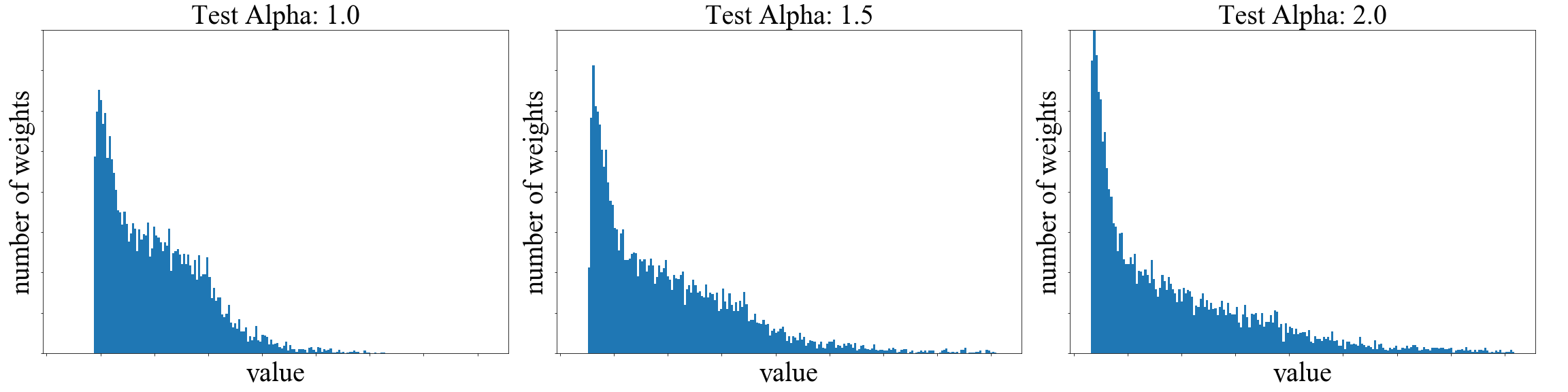}
  \caption{Baseline}
\end{subfigure}
\begin{subfigure}{\textwidth}
  \centering
  \includegraphics[width=0.6\linewidth]{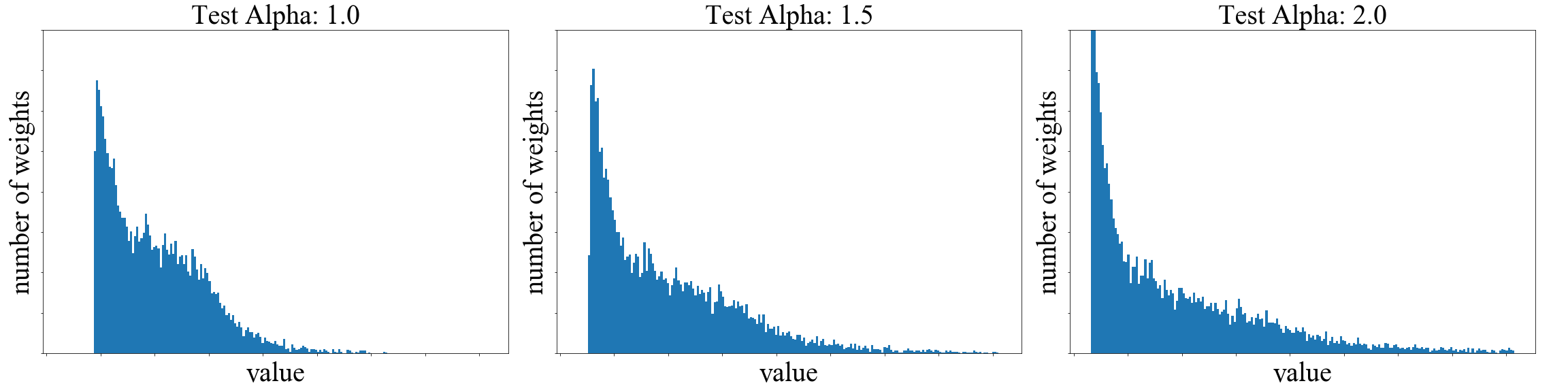}  
  \caption{Training Alpha = 0.1}
\end{subfigure}
\begin{subfigure}{\textwidth}
  \centering
  \includegraphics[width=0.6\linewidth]{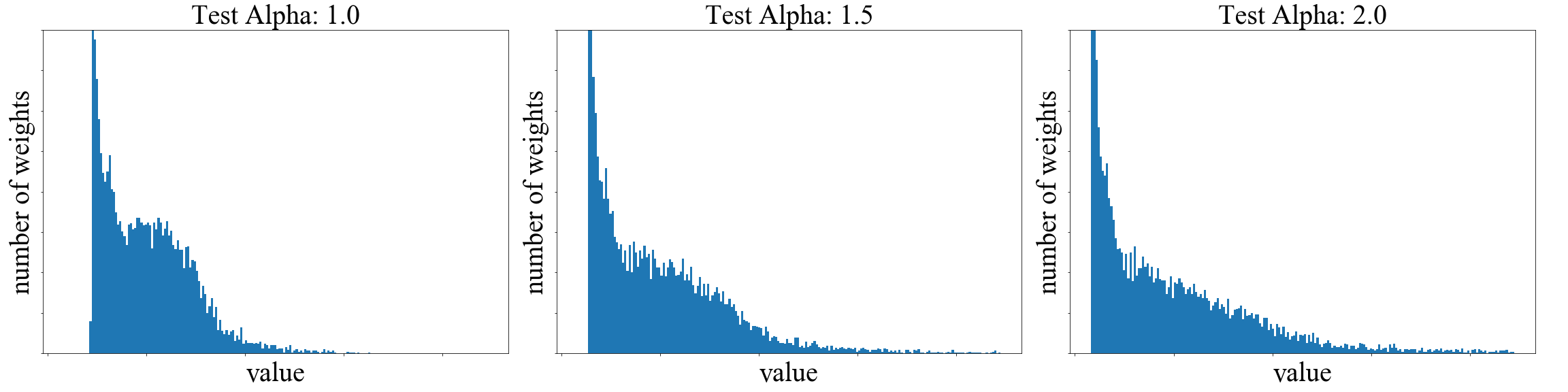}  
  \caption{Training Alpha = 1.0}
\end{subfigure}
\begin{subfigure}{\textwidth}
  \centering
  \includegraphics[width=0.6\linewidth]{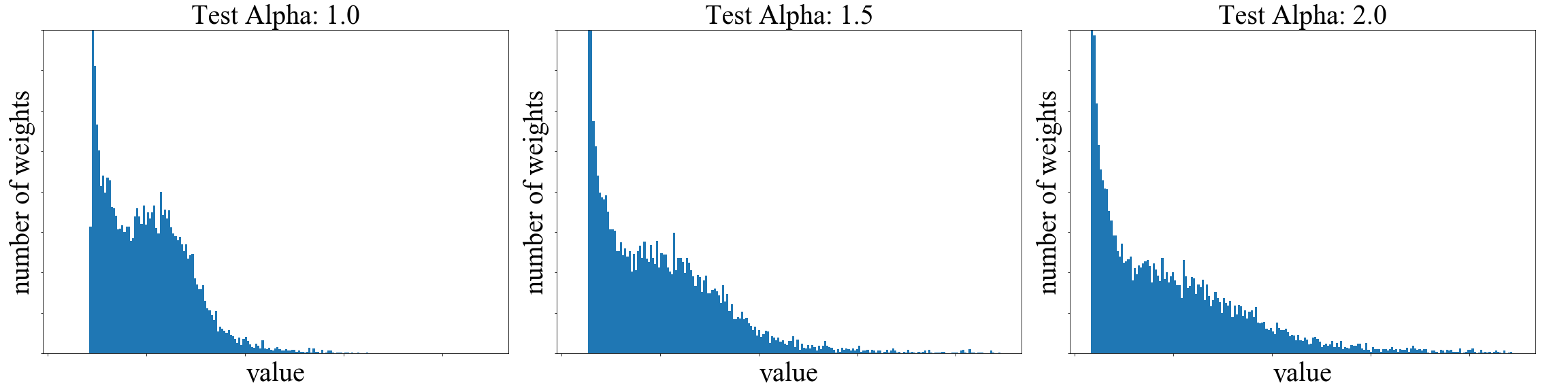}  
  \caption{Training alpha = 1.0}
\end{subfigure}
\begin{subfigure}{\textwidth}
  \centering
  \includegraphics[width=0.6\linewidth]{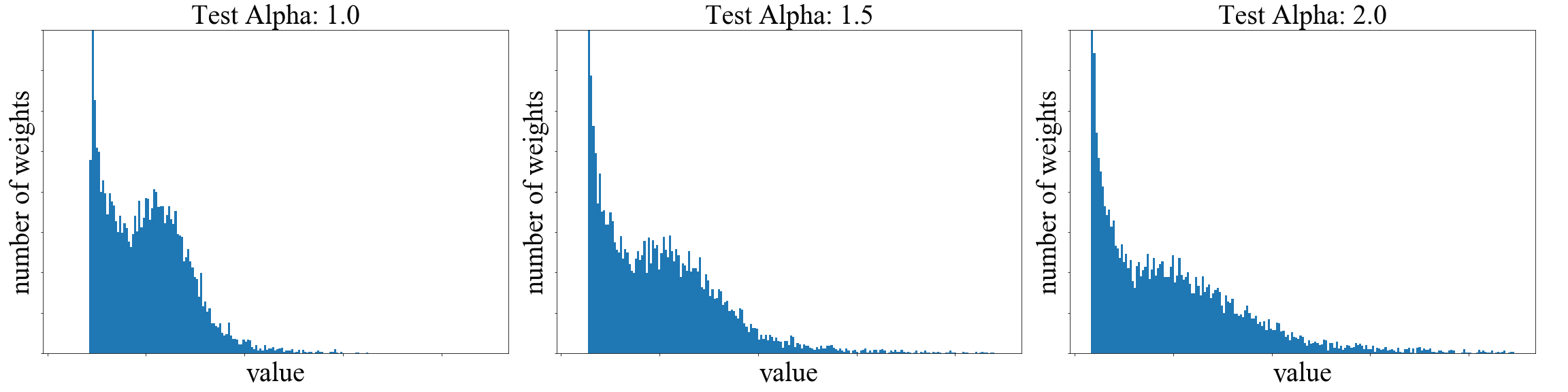}  
  \caption{Training alpha = 2.0}
\end{subfigure}
\caption{We visualize the importance distribution over the entire test set. The weights are evaluated with $\epsilon=0.016$}
\label{fig:fig2}
\end{figure}

\begin{figure}[ht]
\begin{subfigure}{\textwidth}
  \centering
  \includegraphics[width=0.6\linewidth]{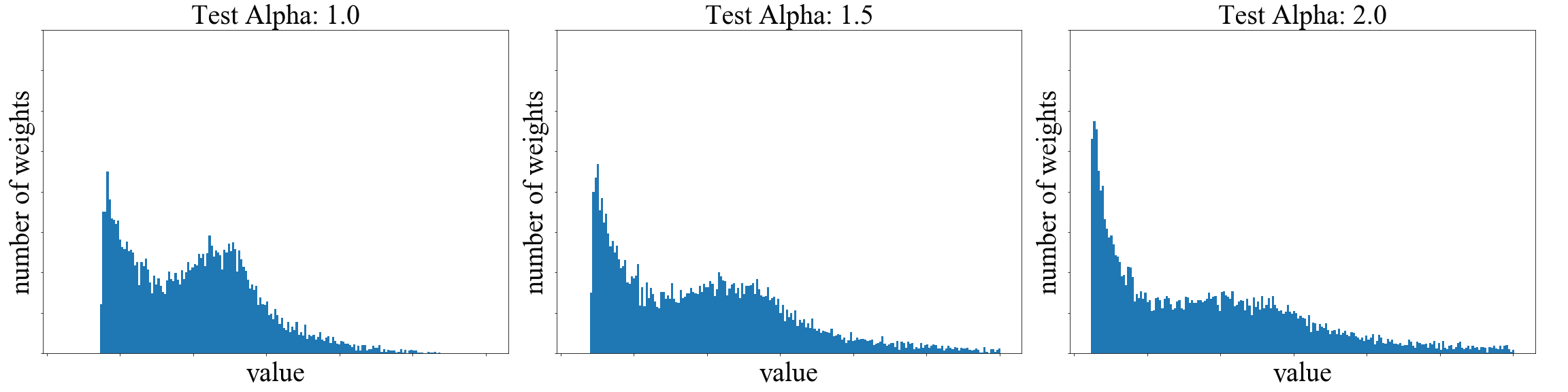}
  \caption{Baseline}
\end{subfigure}
\begin{subfigure}{\textwidth}
  \centering
  \includegraphics[width=0.6\linewidth]{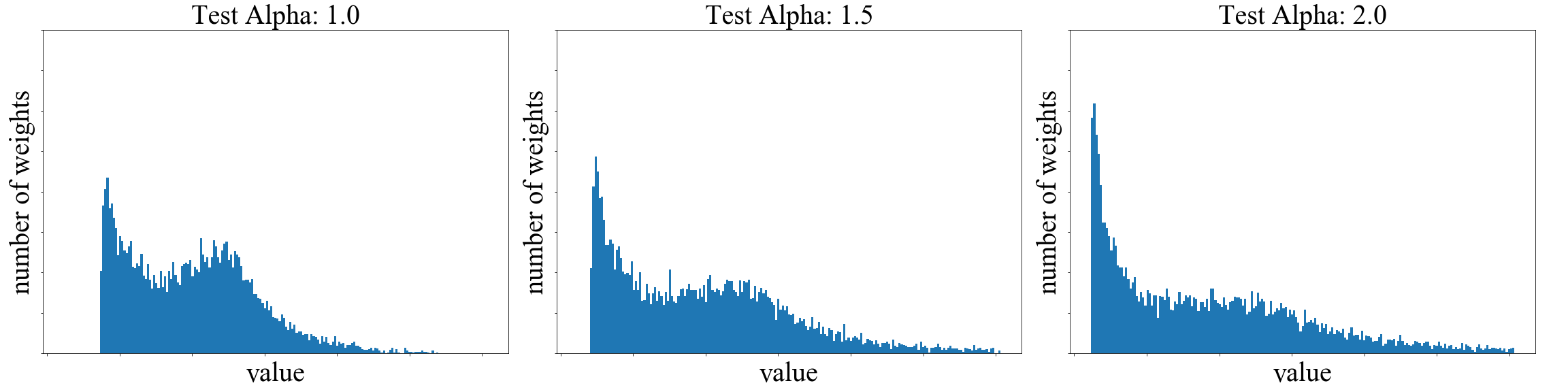}  
  \caption{Training Alpha = 0.1}
\end{subfigure}
\begin{subfigure}{\textwidth}
  \centering
  \includegraphics[width=0.6\linewidth]{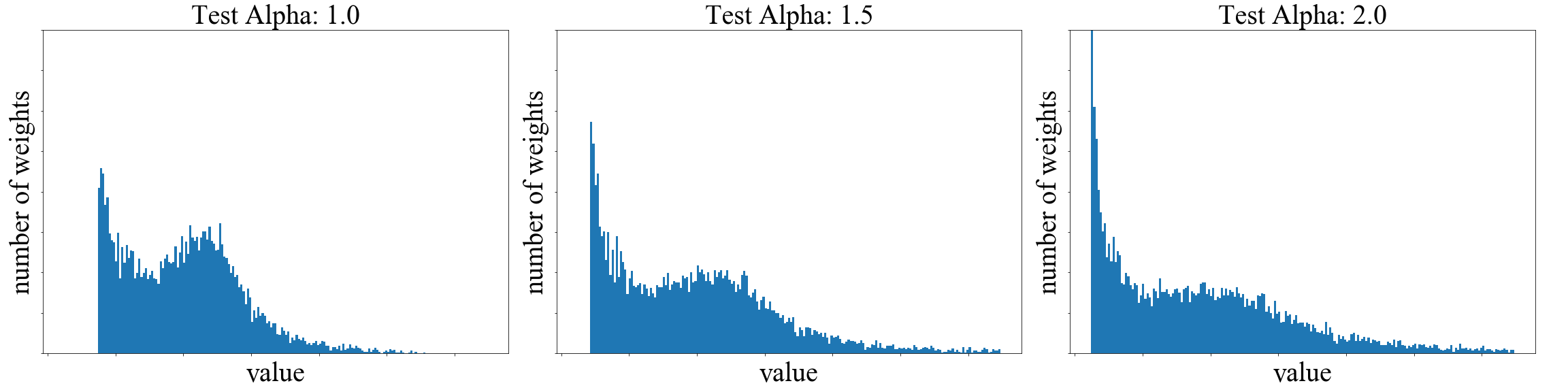}  
  \caption{Training Alpha = 1.0}
\end{subfigure}
\begin{subfigure}{\textwidth}
  \centering
  \includegraphics[width=0.6\linewidth]{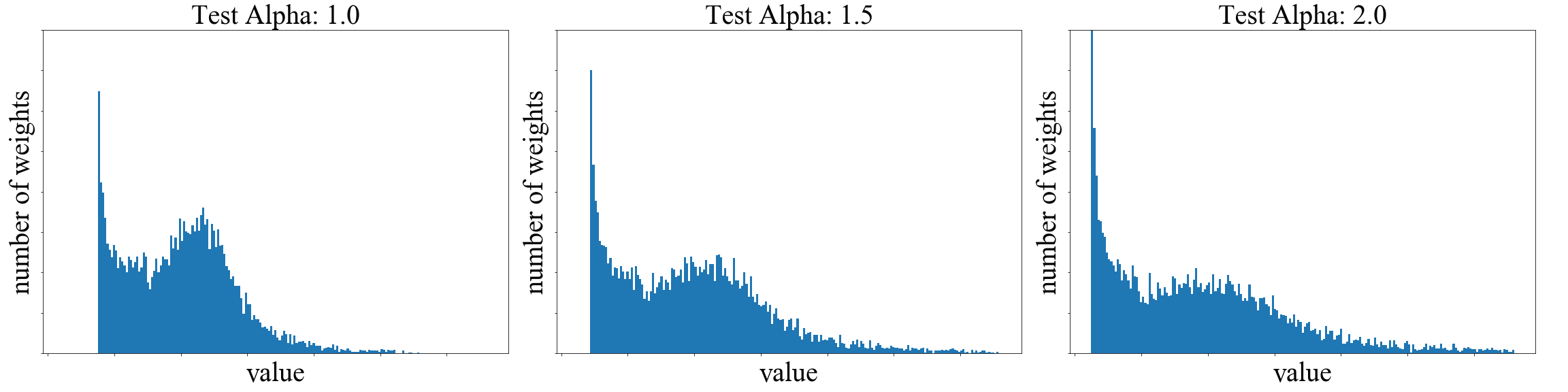}  
  \caption{Training Alpha = 1.5}
\end{subfigure}
\begin{subfigure}{\textwidth}
  \centering
  \includegraphics[width=0.6\linewidth]{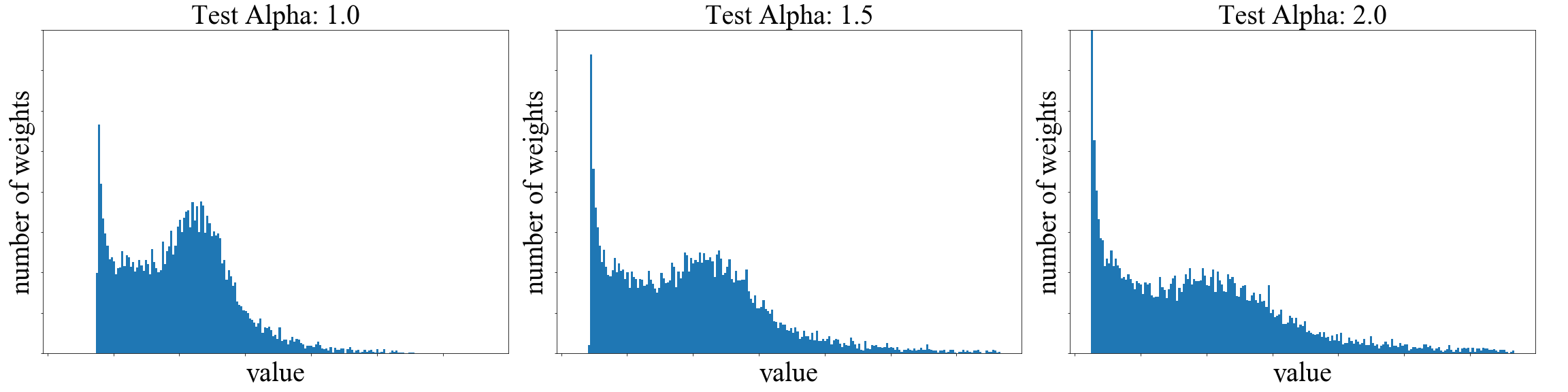}  
  \caption{Training Alpha = 2.0}
\end{subfigure}
\caption{We visualize the importance distribution over the entire test set. The weights are evaluated with $\epsilon=0.031$}
\label{fig:fig3}
\end{figure}

\end{document}